\documentclass{article}
\usepackage{ijcai26}

\usepackage{times}
\usepackage{soul}
\usepackage{url}
\usepackage[hidelinks]{hyperref}
\usepackage[utf8]{inputenc}
\usepackage[small]{caption}
\usepackage{graphicx}
\usepackage{amsmath}
\usepackage{amsthm}
\usepackage{booktabs}
\usepackage{algorithm}
\usepackage{algorithmic}
\usepackage[switch]{lineno}
\usepackage{natbib}
\usepackage{multirow}

\usepackage{bold-extra}
\usepackage{array}
\usepackage{booktabs}
\usepackage[acronym]{glossaries-extra}
\setabbreviationstyle[acronym]{long-short}
\setabbreviationstyle[short]{short-nolong}
\newacronym{CTP}{\textsc{ctp}}{\textit{Canadian Traveler Problem}}
\newacronym{CTPwS}{\textsc{ctp}w\textsc{s}}{\textit{Canadian Traveler Problem with Scouts}}
\newacronym{SAP}{\textsc{sap}}{\textit{Scout-Assisted Planning}}
\newacronym{MCTS}{\textsc{mcts}}{\textit{Monte Carlo Tree Search}}
\newacronym{IAP}{\textsc{iap}}{\textit{Information Gain-based Action Pruning}}
\newacronym{DAP}{\textsc{dap}}{\textit{Distance-based Action Pruning}}
\newacronym{IG}{\textsc{ig}}{\textit{Information-Gain}}
\newacronym{SAPIAP}{\textsc{sap}-\textsc{iap}}{\gls{SAP} \textit{with Information Gain Action Pruning}}
\newacronym{SAPL}{\textsc{sap}-\textsc{liap}}{\gls{SAP} \textit{with Learning-informed Action Pruning}}
\newacronym{SAPD}{\textsc{sap}-\textsc{dap}}{\gls{SAP} \textit{with Distance-based Action Pruning}}

\newacronym{POMCP}{\textsc{pomcp}}{\textit{Partial Observation Monte Carlo Planning}}
\newacronym{GNN}{\textsc{gnn}}{\textit{Graph Neural Network}}
\newacronym{MDP}{\textsc{mdp}}{\textit{Markov Decision Process}}
\newacronym{UAV}{\textsc{uav}}{\textit{Unmanned Aerial Vehicle}}
\newacronym{UGV}{\textsc{ugv}}{\textit{Unmanned Ground Vehicle}}
\newacronym{UCB}{\textsc{ucb}}{\textit{Upper Confidence Bounds}}
\newacronym{POMDP}{\textsc{pomdp}}{\textit{Partial Observation MDP}}
\newacronym{PBP}{\textsc{pbp}}{\textit{Possible Blocking Point}}
\usepackage{xcolor}

\newcommand{\ranenv}{\text{Dense Urban Town}}
\newcommand{\islandenv}{\text{Rural Villages}}
\newcommand{\bridgesenv}{\text{City with river-crossing}}

\newcounter{algoline}

\title{Scout-Assisted Planning for Heterogeneous Robot Teams under Partially Known Environments}

\author{
Hoang-Dung Bui
\and
Abhish Khanal
\and
Raihan Islam Arnob
\and
Gregory J. Stein
\\
\affiliations
George Mason University\\
\emails
\{hbui20, akhanal7, rarnob, gjstein\}@gmu.edu,
}

\begin{document}

\maketitle

\begin{abstract}
Autonomous robot teams navigating partially known environments face costly backtracking when ground robots encounter blocked roads that are only revealed upon physical traversal. We address this with \glsentryfull{SAP}, a heterogeneous planning framework in which scouting \glspl{UAV} proactively gather environmental information to improve \gls{UGV} navigation. To focus scouting on the most consequential edges, we propose \glsentryfull{IAP}, which scores candidate scouting actions by their expected impact on ground robot behavior. Since exact \gls{IAP} computation is prohibitively expensive, we develop a \gls{GNN}-based model that predicts information gain values directly from graph structure and belief state, reducing planning time to real-time levels without sacrificing solution quality. Experiments across three environment types show that \gls{SAPIAP} reduces ground robot travel cost by 31.9--37.7\% over the \gls{CTP} baseline, and outperforms proximity-based scouting guidance by an additional 8--14\%, confirming that principled information-gain-guided scouting is both more effective and computationally feasible for real-world deployment.
\end{abstract}

\section{Introduction} 
\label{sec:Intro}
Autonomous robot teams operating in disaster-struck or hazardous environments must navigate through partially known terrain where roads or pathways may be blocked by debris, flooding, or structural damage. Making efficient decisions under this uncertainty is critical — unnecessary detours and backtracking waste time and resources that could be life-saving in search-and-rescue or logistics scenarios.

This problem is formalized as the \gls{CTP}, originally introduced by Papadimitriou and Yannakakis~\cite{papadimitriou1991shortest}. In \gls{CTP}, a \gls{UGV} navigates in a graph where some edges may be blocked, and blockages are only revealed upon physical traversal. When a robot reaches a blocked edge, it must backtrack and replan — a costly cycle that grows more expensive as environments scale in complexity.

\begin{figure}
    \centering
    \setlength{\tabcolsep}{-1pt}
    \begin{tabular}{cc}
        \includegraphics[width=0.50\linewidth]{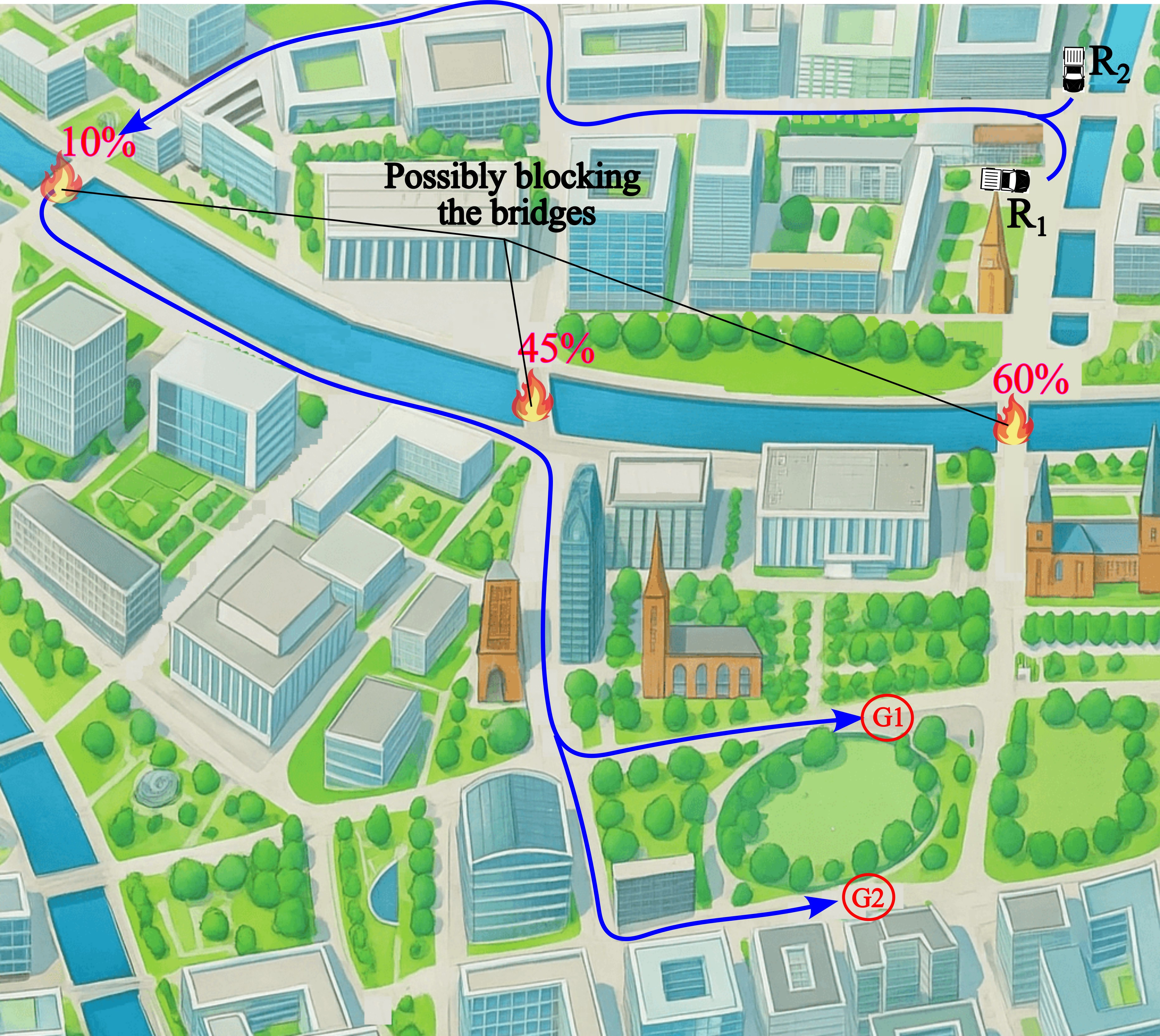} & 
        \includegraphics[width=0.5\linewidth]{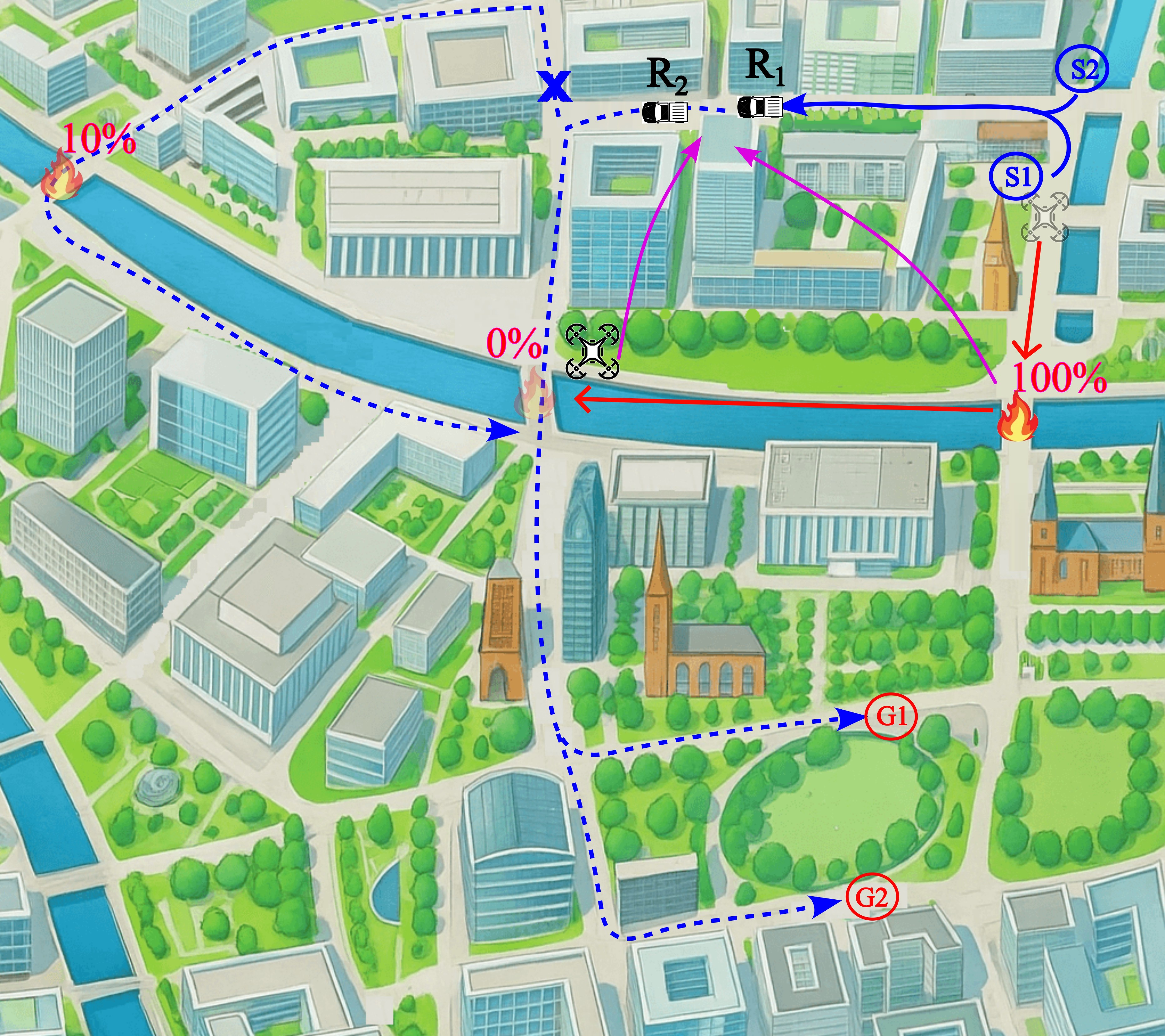} \\
        \footnotesize{a) Canadian Traveler Problem (\textsc{ctp})} & 
        \footnotesize{b) Scout-Assist Planning (\textsc{sap})}
    \end{tabular}
    \caption{\textbf{Gaining critical environmental data earlier improves significantly travel distance for \gls{UGV}'s navigation}. In \gls{CTP} (a), a \gls{UGV} follows its policy until facing a blocked street, then replan. Gaining environmental information sooner from a scouting \gls{UAV} (b), the \glspl{UGV} can replan and change its behavior earlier that saves significant travel cost.}
    \label{fig:frontpage}
\end{figure}

A natural remedy is to augment the ground team with scouting \glspl{UAV}. Unlike ground robots, \glspl{UAV} are unconstrained by terrain and can rapidly overfly edges to verify their status, feeding that information back to improve ground robot planning. However, deploying scouts can be inefficient without principled guidance. Without a strategy for choosing which edges to inspect, drones may expend resource on low-utility edges that are irrelevant to the ground team's current path — providing information that never changes robot behavior. Computing an optimal scouting policy by sampling over all unvisited edges is computationally intractable due to the exponential branching factor this introduces.

A straightforward heuristic is to guide drones toward edges spatially close to the ground robots — a proximity-based strategy that ensures scouted information is locally relevant. This reduces unnecessary backtracking and lowers ground robot travel cost compared to unguided scouting (up to 26\% in our experiments). However, proximity-based guidance is inherently myopic: it focuses drones on nearby edges regardless of whether those edges are actually on any plausible future path for the ground team, ignoring the long-term structure of the planning problem.

To move beyond local heuristics, we formulate an \gls{IG} criterion that estimates the expected reduction in ground robot travel cost if a given edge's status were known. By ranking edges according to this value and pruning low-utility candidates, the drone's attention is concentrated on edges that meaningfully alter the ground team's decisions. This yields further improvements over the distance-based heuristic (around 12\% in our evaluation). However, computing \gls{IG} values exactly requires evaluating ground robot policies under all possible edge-status combinations — a process too computationally expensive for real-time deployment.

To close the gap between solution quality and computational feasibility, we develop a \gls{GNN} model trained to predict \gls{IG} values directly from the graph structure and current robot state. The \gls{GNN} captures the relational dependencies between edges — which paths the ground robot is likely to use, which blockages would force costly detours — and produces accurate \gls{IG} estimates in a fraction of the time required by exact computation. This makes the full planning framework applicable to real-time settings.

We formalize this setting as the \gls{CTPwS}, a new variant of \gls{CTP} in which a heterogeneous team of \glspl{UGV} and \glspl{UAV} collaboratively navigates a partially known environment toward a shared goal. Although the map topology is known, edge blockages due to weather or disaster events are initially unknown and only verifiable by direct inspection. \gls{CTPwS} inherits the exponential state-space growth of \gls{CTP} — compounded by the drone's unconstrained action space across all unvisited edges — and additionally requires a unified high-level action representation applicable to both \glspl{UGV} and \glspl{UAV}.

Existing work on scout-assisted planning~\citep{stadler2023approximating} partially addresses \gls{CTPwS} through a hybrid approach that computes a joint policy only when scouting is expected to improve overall performance. This approach relies on an approximated \gls{IG} formulation that assume a deterministic ground robot finishing time — an assumption that breaks down with more than one ground robot, restricting the framework to single-\gls{UGV} settings.

We present \glsentryfull{SAPIAP}, a learning-informed planning framework for heterogeneous robot teams that is tractable at scale. Our contributions are:
\begin{itemize}
\item An \glsentryfull{IG} formulation that estimates the value of each \gls{UAV} scouting action and focuses exploration on a compact, high-utility subset of edges.
\item A \gls{GNN}-\textit{based predictor} that approximates \gls{IG} values in real time, reducing planning overhead from seconds to milliseconds and enabling deployment in dynamic environments.
\item \textit{Empirical evaluation} on three types of environments demonstrating up to (31.5--51.2)\% reduction in ground robot travel cost compared to baselines.
\end{itemize}

\section{Related Work}
\label{sec:Relw}

\paragraph{\glsentryfull{CTP}}
\gls{CTP} was introduced by \citet{papadimitriou1991shortest} and characterized as a deterministic \gls{POMDP} and P-hard problem. Three broad families of approaches have been proposed to address it.
The first family adopts an \textit{optimistic assumption}: all unknown edges are treated as passable, and the agent computes a shortest path on this free-space graph. When a blocked edge is encountered during execution, the agent replans. \citet{bnaya2009canadian} augmented this strategy with remote sensing, triggering observations only when the expected improvement in travel cost exceeded the sensing overhead. \citet{eyerich2010high} contributed two sampling-based algorithms within this family: \textit{Hindsight Optimization} and \textit{Optimistic Rollout}, which differ in how rollout costs are estimated. Hindsight Optimization uses the shortest path on a sampled graph, while Optimistic Rollout accumulates travel cost along optimistic shortest paths until a blocked edge is encountered or the goal is reached. While computationally efficient, optimistic methods force the agent to replan repeatedly upon encountering blocked edges, causing travel costs to accumulate significantly over time.
The second family incorporates edge uncertainty directly into the search. \citet{nikolova2008route} proposed a \textit{minimum expected distance} criterion to estimate edge costs under blocking probabilities, paired with an \textit{expected minimum distance} heuristic to approximate cost-to-goal. \citet{guo2019robust} extended this line with \textsc{LAO}* search, which trades off policy quality against planning time by minimizing \textit{Exponential Risk} — the variance from the optimal expected cost. More recently, \citet{veys2024generating} reduced runtime by constructing sparse probabilistic graphs that prune edges unlikely to improve the expected plan cost, narrowing the search space without sacrificing solution quality.
The third family reasons over expected shortest paths on the uncertain graph directly. \citet{narayanan2017heuristic} computed all relevant shortest paths along with their probabilities and selected the path with the lowest expected cost. \citet{bampis2022canadian} proposed a CAO* algorithm with a caching mechanism to avoid re-expanding nodes in the AO* tree, combined with an upper-bound heuristic to preserve admissibility. Despite these improvements, both approaches require polynomial runtime in the number of edges, limiting their scalability to larger graphs. \citet{eyerich2010high} also presented a \gls{UCB}-based algorithm that uses sampling with agent history to evolve belief states, computing rollout costs as the sum of travel cost with a heuristic evaluated on the optimistic graph. We build on this idea in developing our own planner for \gls{CTP}.
Across all three families, a shared limitation is that the robot must physically traverse an edge to verify its status — meaning blocked edges are discovered only after costly travel. This motivates integrating scouting agents that can gather information proactively, before the ground robot commits to a path.
\paragraph{\glsentryfull{CTPwS}}
Several frameworks have explored leveraging scouting robots to improve the planning of other team members, though none fully addresses the challenges of the \gls{CTPwS} setting.
\citet{stadler2023approximating} propose the most closely related work, restricting attention to a single ground robot supported by multiple sensing agents. Their approach uses deterministic navigation macro-actions defined over the observed graph, and derives policy approximation formulas under the assumption that the ground robot is the sole source of stochasticity. This assumption becomes untenable when multiple ground robots are introduced: each robot can trigger independent belief updates, causing macro-action termination times to become unpredictable under the abstraction used. As a result, the policy approximation formulas are invalidated, and the framework does not extend to teams with more than one ground robot. Addressing this limitation requires joint reasoning over the interruption events that arise from shared belief updates across all agents.

\paragraph{Graph Neural Networks for Planning under Uncertainty}
GNNs have recently emerged as a powerful tool for learning heuristics and value estimates over graph-structured problems, replacing hand-crafted rules with strategies inferred directly from data~\citep{drori2020learning}. In robotics, this paradigm has been extended to planning under uncertainty: \citet{chen2020autonomous} combine GNNs with deep reinforcement learning to predict optimal sensing actions in belief space for autonomous exploration, while \citet{anorb2023aig} use a GNN to reason over both local and non-local graph structure to improve navigation under partial observability. However, these approaches either target fully observable routing problems or learn end-to-end action policies in continuous exploration settings. None address the problem of predicting per-edge information gain values in a partially observable graph where an edge's scouting utility depends jointly on the ground robot's belief state, its planned path, and the probabilistic structure of remaining unknown edges.

In our work, we address this gap by building on \gls{POMCP} \citep{silver2010monte} to solve the classical \gls{CTP}, following the spirit of the \gls{UCB}-based approach of \citet{eyerich2010high}. Like that method, \gls{POMCP} is a sampling-based planner that constructs a Monte Carlo tree to evolve belief states from agent history. From this foundation, we develop \gls{SAP} to solve \gls{CTPwS} for a heterogeneous robot team on uncertain graphs, supporting multiple ground robots without the single-robot restriction of prior work.

\section{Problem Formulation}
\label{sec:Prob}
We consider a heterogeneous robot team consisting of two sub-teams: a ground team of $N$ \glspl{UGV} and a scout team of $M$ \glspl{UAV}, operating in a partially known environment. 
The ground team is tasked to reach $N$ designated goal locations, while the scout team is assigned to rapidly gather environmental information to improve the ground team's navigation. As new information becomes available, the ground team updates its behavior to benefit from the incoming observations.

Environments are represented as graphs in which edges and vertices correspond to roads and intersections, respectively.
Due to weather or disaster events, roads might be blocked at specific points. To verify whether a road is blocked, a robot must physically visit the corresponding point. We assume that each edge $e_i$ has one \gls{PBP}, denoted as \textsc{b}$_i$, with blocking probability $p_i$, located at the midpoint of the edge.
\glspl{PBP} are treated as additional vertices in the graph.
Accordingly, a graph \textsc{g} in our formulation consists of two types of nodes: regular vertices (intersections) with zero blocking probability and \glspl{PBP}.
This unified graph representation enables a share high-level action space for both ground robots and scouting drones.
We further assume that observations are perfect and instantaneous relative to travel time, so observation cost is negledted. 

\paragraph{Defining Abstract Action}
\label{sec:action_def}
Unlike prior \gls{CTP} work in which actions begin and end at vertices, our high-level action $\sigma$ must accumulate arbitrary robot locations since robots immediately interrupt and restart their actions upon receiving updated environmental information. Accordingly, each action $\sigma$ originates from the robot's current location and navigates to a graph's vertex or a \gls{PBP} $\textsc{b}_i$.
Each action also contains a robot's identifier, ensuring correct assignment when multiple robots complete actions simultaneously.

\begin{figure}
    \centering
    \includegraphics[width=0.7\linewidth]{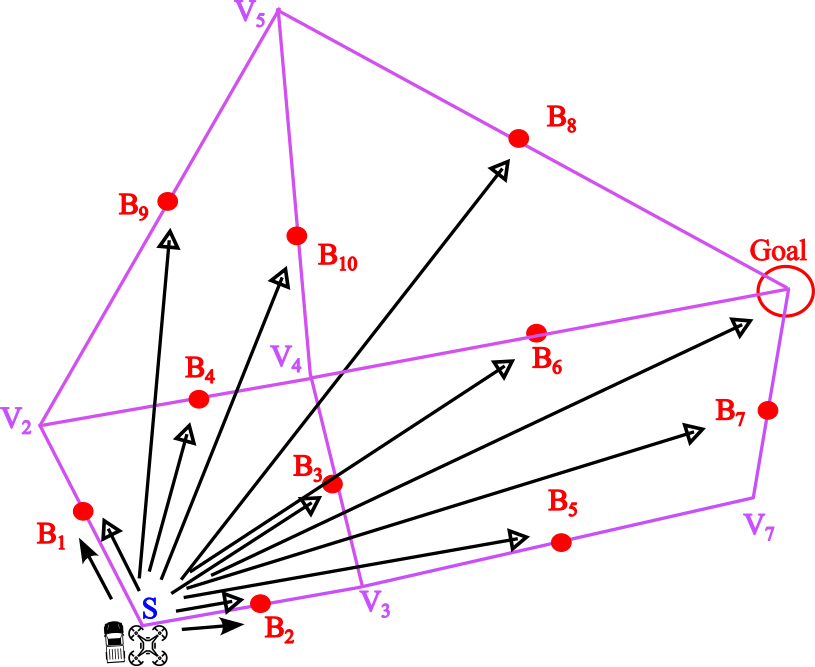}
    \caption{\textbf{Abstract Actions for both \glspl{UAV} \glspl{UGV}}. The \gls{UGV} has two actions: move to \{$\textsc{b}_1, \textsc{b}_2 $\}, while the \gls{UAV} has a action set of all unvisited \glspl{PBP} \{$\textsc{b}_1,\textsc{b}_2,\ldots,\textsc{b}_{10}$\}.}
    \label{fig:actions}
\end{figure}

\begin{figure}[ht]
    \centering
    \includegraphics[width=0.95\linewidth]{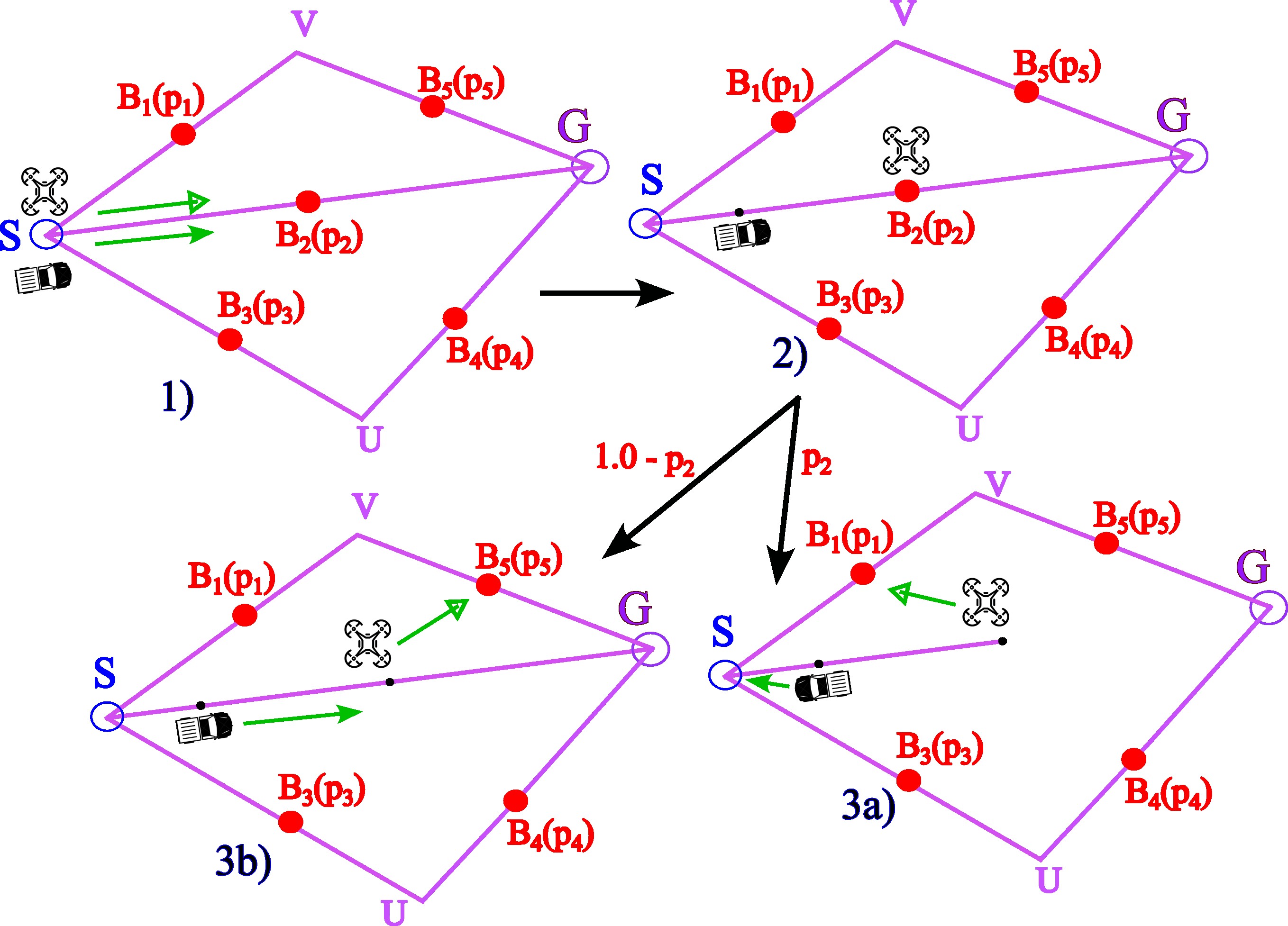}
    \caption{\textbf{Transition Model of \glsentryfull{SAP} for a team of 1 \gls{UGV} \& 1 \gls{UAV}} . The \gls{UGV} is tasked to reach its goal \textsc{g} with the minimal expected cost.
    The graph has four vertices: $V = \{\textsc{s}, \textsc{g}, \textsc{u}, \textsc{v}\}$ and five edges with five  \glspl{PBP}$ = \{\textsc{b}_1,\textsc{b}_2, \textsc{b}_3, \textsc{b}_4, \textsc{b}_5\}$.  
    At step 1, both robots are assigned to move to $\textsc{b}_2$. 
    The drone reaches its goal first (step 2). The observation of $\textsc{b}_2$ resets the action of the ground robot. 
    If $\textsc{b}_2$ is blocked ($p_2$ percent), the \gls{UGV} should terminate its current action and go back to $\textsc{s}$ (step 3a). The \gls{UAV} then heads to $\textsc{b}_1$ to support the ground robot making decision as reaching \textsc{s}.
    At step 3b, there is (1-$p_2$) percent $\textsc{b}_2$ is traversable, and the ground robot continues its action. The information from the drone is not matter to the ground robots from this point.
    The process is continue until the robots reach their goals or the sampling tree gets its maximum depth.}
    \label{fig:transition_model}
\end{figure}

\paragraph{Defining Belief State}
The belief state $b_t$ in \gls{CTPwS} is discrete with respect to each robot’s binary observations of whether a given road is blocked or not.
We assume lossless communication between all robots, so every agent maintains an identical, up-to-date partial graph at all times.

Formally, the belief state at time $t$ is defined as
$b_t(s) = P(s_t=s|h_t)$, where $h_t$ is the history of actions, observations, and initial belief, and $s_t$ is the true state at time $t$. The true state is represented as the tuple:

\begin{equation}
    b_t = \{p_g, p_d, \mathcal{G}, \mathcal{B}_U, \mathcal{B}_T, \mathcal{B}_B\}
\end{equation}
where $p_g$, $p_d$ are the current positions of ground robots and drones, respectively; $\mathcal{G}$ is a graph representing the environment; and $\mathcal{B}_U$, $\mathcal{B}_T$, $\mathcal{B}_B$ are sets of unknown, traversable, and blocked points, respectively.

\paragraph{POMDP Formulation}
Planning is centrally coordinated: at each time step, a joint action $a_t$ is computed for the entire team, specifying each robot's next movement so as to minimize the expected travel cost of the ground team.
Formally, the problem is cast as a \glsentryfull{POMDP}~\citep{kaelbling1998planning}. The expected cost $Q$ under this model is expressed via the belief-space Bellman Equation~\citep{pineau2002integrated} : 

\begin{align}
    Q(b_t, a_t) = \sum_{b_{t+1}} P(b_{t+1} |b_T,a_t) [R(b_{t+1}, b_t, a_t) \nonumber \\ + \underset{{a_{t+1}\in A(b_{t+1})}}{\text{min}} Q(b_{t+1}, a_{t+1})]
\end{align}
where $R(b_{t+1}, b_T, a_t)$ is the cost accumulated by reaching belief state $b_{t+1}$ from $b_t$ by taking action $a_t$.
The objective is to find a policy $\pi$ that minimizes the expected cumulative cost of the ground robot team.

\section{Scout-Assisted Planning (SAP)}
We present \glsentryfull{SAP}, a model-based planning framework for heterogeneous robot teams operating in partially known environments. 
\gls{SAP} enables concurrent action execution across ground robots and scouting drones, allowing the team to share observations in real time and update behavior as new environmental information arrives.
Building on the multi-robot sub-goal planning framework of \cite{khanal2023learning}, in which robot actions are concurrent and may finish at different times, we extend the transition model to support heterogeneous robot teams composed of both \glspl{UGV} and and \glspl{UAV}.

The key distinction of our model is that robots may \textit{terminate} their current actions immediately upon receiving new observations, enabling ground robots to replan and adapt as the scout team resolves edge uncertainty.
This capability introduces two modeling changes: (1) robots may interrupt and restart actions from arbitrary positions rather than fixed graph's vertices, requiring a new high-level action definition that is not anchored to graph vertices, and (2) the action spaces of drones and ground robots are structurally different, requiring a hierarchical treatment of joint actions. 

In this section, we formalize the \gls{SAP} state representation,  the concurrent state transition model, and the joint-action Bellman equation used to compute expected team cost.
We then present a \gls{DAP} heuristic that prunes low-value drone actions by prioritizing \glspl{PBP} near the ground team, yielding a significant reduction in travel cost over unguided scouting. 
Building on its limitations, we introduce \glsentryfull{IAP}, a principled pruning strategy that guides scouting drones toward edges with highest expected impact on ground robot behavior. Finally, we describe how \gls{IAP} is integrated with \gls{SAP} into the full \gls{SAPIAP} planner.

\subsection{Expected Cost of Scout-Assisted Planning with High-Level Joint Actions}
A high-level joint action $a_t$ is a list of single-robot actions $a_t = [\sigma_1,\ldots, \sigma_N, \sigma_{N+1}, \ldots, \sigma_{N+M}]$, assigning each robot a target vertex or a \gls{PBP} to navigate to. For a belief state $b_t$, the joint action set $A(b_t)$ is the outer product of all individual robot's action sets, where \glspl{UGV} may move to neighboring vertices or \glspl{PBP} along edges, and \glspl{UAV} may fly directly to any unvisited \glspl{PBP}. To avoid redundant scouting, no two drones can be assigned the same \gls{PBP} simultaneously. 

The action sets of drones and ground robots differ in an important structural way: Drones share a common action set consisting of all unvisited \glspl{PBP}, which shrinks monotonically as scouting progresses. Ground robots, by contrast, act on their local neighborhood of vertices, and their action sets remain stable in size through execution. 

\paragraph{State Transition.} 
The core principle of the \gls{SAP} transition model is that whenever any robot acquires a new observation, all other robot immediately terminate their current actions and are reassigned based on the update belief state.
If a robot completes its action without generating a new observation, the planner simply assigns it a new action.
The high-level satate is thus updated at the moment of the first robot among the team completes its assigned action.
The time cost of a joint action $T'$ is therefore determined by the robot completing its action first.

\begin{align}
    T'(b_t,a_t) = \underset{\forall \sigma \in a_t}{\text{min}} (T(b_t, \sigma)); \ 
    \sigma'(b_t, a_t) = \underset{\forall \sigma \in a_t}{\text{argmin}} (T(b_t, \sigma)) 
\end{align}


The outcome of completing joint action $a_t$ depends on whether the completing robot reaches an unvisited \gls{PBP}. If so, the observation is binary: the \gls{PBP} is blocked with probability $P_B(\sigma')$, yielding successor state $b_B$, or traversable with probability $1 - P_B(\sigma')$, yielding successor state $b_T$:

\begin{align}
    b_B = \langle p^t (a_t,T'), \textsc{g}, & \textsc{b}_U' = \textsc{b}_U \backslash \{ \sigma'\}, \textsc{b}_B' = \textsc{b}_B \cup \{ \sigma'\}, \nonumber \\
    & \textsc{b}_T' = \textsc{b}_T \rangle \\
    b_T = \langle p^t (a_t,T'), \textsc{g}, & \textsc{b}_U' = \textsc{b}_U \backslash \{ \sigma'\}, \textsc{b}_B' = \textsc{b}_B, \nonumber \\
    & \textsc{b}_T' = \textsc{b}_T \cup \{ \sigma'\} \rangle \nonumber
\end{align}

If the completing robot instead reaches or graph vertex or an already-known \gls{PBP}, no new observation is generated and only the robot positions are updated:

\begin{align}
    b_N = \langle p^t (a_t,T'), \textsc{g}, \textsc{b}_U' = \textsc{b}_U, \textsc{b}_B' = \textsc{b}_B, \textsc{b}_T' = \textsc{b}_T \rangle
\end{align}

\paragraph{Bellman Equation.}
Given our state and action abstraction above, the expected cost of a joint action $a_t$ in belief state $b_t$ is defined by the following Bellman equation (see also Fig.~\ref{fig:transition_model}).

\begin{align}
\label{eq:bell_joint}
    Q(b_t, \sigma_t \in & A(b_t)) = T' + P_B(\sigma') \underset{a_{t+1}\in A'(b_B)}{min} Q(b_B, a_{t+1}) \nonumber \\
    & + [1-P_B(\sigma')]\underset{a_{t+1}\in A'(b_T)}{min} Q(b_T, a_{t+1})
\end{align}

This equation accumulates the travel cost $T'$ of the current join action and branches over the two possible observation outcomes, each leading to a recursively evaluated successor state.
The \gls{SAP} framework---comprising the state representation, joint action formulation, and transition model---is solved approximately using \gls{POMDP}~\citep{silver2010monte}, a sampling-based solver that constructs a Monte Carlo tree over belief states without requiring explicit state enumeration.

\subsection{Distance-Based Action Pruning (\textsc{dap})}
\label{sec:dap}

While \gls{SAP} provides a complete planning framework for the heterogeneous team, the drone's action set grows with the number of unvisited \glspl{PBP}, making exhaustive evaluation of all scouting candidates computationally expensive.
A natural first approach to guide the scouting drone is to prioritize \glspl{PBP} that are spatially close to the ground robots. The intuition is straightforward: edges near the ground team are more likely to be on their imminent path, so resolving their status early reduces the chance that a robot commits to a blocked route.
Formally, the priority score for drone $j$ to scout \gls{PBP} $\textsc{b}_e$ under the distance-based heuristic is:
\begin{equation}
\label{eq:dist_heuristic}
v^j_D(b, a_e) = \frac{1}{\sum_{i=1}^N \text{dist}(q_{g_i},\ q_{b_e})}
\end{equation}
where $q_{g_i}$ is the current position of ground robot $i$ and $q_{b_e}$ is the location of \gls{PBP} $\textsc{b}_e$. 
The drone is assigned to the highest-scoring candidate, prioritizing the \glspl{PBP} that minimizes the distance to the ground robot team.

While this heuristic is computationally lightweight and improves over unguided scouting by directing the drone toward locally relevant edges, it is inherently myopic. Proximity to the ground robot team does not imply that an edge lies on any plausible future path toward the team's goals — a nearby edge that is structurally irrelevant to all ground robot plans will score highly under $v^j_D$ while displacing scouting effort from edges that would genuinely alter robot behavior. This limitation motivates a more principled formulation that accounts for the global structure of the planning problem, which we develop next.

\subsection{Information Gain-based Action Pruning (\textsc{iap})}

To address the limitation of \gls{DAP}, we introduce \glsentryfull{IAP}, a principled pruning strategy that focuses the drone's attention on edges whose observation is most likely to improve ground robot behavior.

\paragraph{Value Change of a Scouting Action.}
Consider a belief state $b$ in which each uncertain edge $e$ carries a blocking probability $p_e in \in (0, 1)$ and contains a \gls{PBP} $\textsc{b}_e$. A scouting action $a_e$ dispatches the drone to observe $\textsc{b}_e$, revealing whether edge $e$ is traversable or blocked. The value of this observation to ground robot $i$ is measured by the change in its value function across the two possible outcomes:
\begin{equation}
\label{eq:vc}
v_c^i(b, a_e) = V^i(b \mid a_e = \text{block}) - V^i(b \mid a_e = \text{trav})
\end{equation}
A large $v_c^i$ indicates that the two outcomes lead to substantially different plans for robot $i$, making edge $e$ a high-priority scouting target. Conversely, a small $v_c^i$ indicates that the observation would not meaningfully alter the robot's behavior regardless of the outcome.

\paragraph{Team-Level Value Change.}
To account for the full ground team of $N$ robots, individual value changes are aggregated:
\begin{equation}
\label{eq:team_vc}
\sum_{i=1}^{N} v_c^i(b, a_e)
\end{equation}

This decoupled formulation is deliberately scalable: each robot's value change is evaluated independently given its start and goal configuration, and contributions are summed at runtime. Adding further ground robots requires no modification to the formulation for existing team members.

\paragraph{Travel Cost.}
The second factor governing action priority is the time required for drone $j$ to reach $\textsc{b}_e$ from its current position:

\begin{equation}
\label{eq:travel_cost}
t_{a_e}^j = \frac{\text{dist}(q_{d_j},\ q_{b_e})}{v_d}
\end{equation}

where $q_{d_j}$ is the drone's current position, $q_{b_e}$ is the location of $\textsc{b}_e$, and $\text{v}_d$ is the drone's velocity. This term penalizes distant scouting targets, reflecting the practical constraint that a delayed observation reaches the ground team too late to meaningfully influence their planning.

\paragraph{IAP Formulation.}
Combining Equations~\ref{eq:vc}--\ref{eq:travel_cost}, the full \gls{IAP} priority value for drone $j$ scouting edge $e$ is:
\begin{equation}
\label{eq:iap}
v^j_I(b, a_e) = \frac{1}{t^j_{a_e}} \cdot p_e(1 - p_e) \cdot \sum_{i=1}^{N} v_c^i(b, a_e)
\end{equation}

The middle factor $p_e(1 - p_e)$ is the variance of the Bernoulli distribution over edge $e$'s blocking state. It is maximized at $p_e=0.5$, where uncertainty is greatest, and vanishes at $p_e \in \{0,1\}$, where the edge state is already known with certainty. This term acts as a natural gating mechanism, suppressing scouting actions on edges that are already effectively resolved regardless of their potential value change. Together, the three factors reward scouting actions that are nearby, target highly uncertain \glspl{PBP}, and are consequential to the ground team's plans---capturing the essential properties of an effective scouting decision in a single interpretable expression.

The travel cost and uncertainty terms are computed analytically from known quantities at runtime. The value change term $\sum_{i=1}^{N} v_c^i$, however, depends on the full graph topology and belief state, and is the central quantity to be estimated. We next describe how this quantity is approximated efficiently using a learned model.

\subsection{\textsc{SAPIAP}: Integrating IAP into SAP}
The complete \gls{SAPIAP} planner integrates the two components described above. At each planning step, \gls{IAP} evaluates the priority score $v^j_I(b, a_e)$ for every candidate scouting action and retains only those exceeding a threshold, reducing the effective branching factor of the drone's action space. The pruned joint action set is then passed to the \gls{POMCP}-based \gls{SAP} solver, which constructs a Monte Carlo search tree over the reduced space to compute the expected cost of each joint action and select the team's next move.

This integration preserves the theoretical guarantees of \gls{POMCP} — asymptotic convergence to the optimal policy as sample count grows — while substantially reducing the per-step computational cost by eliminating low-value scouting candidates before the tree search begins. The practical effect is a planner that scales to realistic graph sizes and team configurations without sacrificing solution quality, as we demonstrate in the experimental evaluation that follows.

\section{Learning-informed Drone's Action Pruning}

Computing the team value change $\sum_{i=1}^{N} v_c^i(b, a_e)$ (Equation~\ref{eq:team_vc}) exactly at runtime requires Monte Carlo sampling over the full graph and belief state at every decision step --- a process that is computationally prohibitive in time-critical scenarios. To address this, we propose a learning-based approach that approximates $v_c^i(b, a_e)$ directly from the graph structure and belief state. A model is trained offline to predict the value change for each edge $e$ and each ground robot $i$ given its start and goal configuration. At runtime, the predicted values are summed across all robots and combined with the analytically computed travel cost and uncertainty terms from Equation~\ref{eq:iap} to produce the final \gls{IAP} priority scores.

\subsection{GNN Structure}

The value change $v_c^i(b, a_e)$ depends on the global structure of the 
graph: an edge that lies on every shortest path between a robot's start and 
goal is far more consequential to scout than one that lies on no shortest 
path. Capturing this dependency requires reasoning over the entire graph 
topology simultaneously, which motivates the use of a \gls{GNN}.

We adopt a \textsc{gat}v2~\cite{brody2022how} architecture, which 
extends the original Graph Attention Network by computing dynamic, 
content-dependent attention coefficients for each edge. This is particularly 
suited to our setting because the relevance of a neighboring node's 
information depends not only on the node's features but also on the edge 
connecting them --- specifically its distance and blocking probability.

\subsubsection{Input Features}

Each node $u$ in the graph is assigned a binary feature vector 
$\mathbf{x}_u \in R^2$ encoding its role for a given ground robot:

\begin{equation}
    \mathbf{x}_u = 
    \begin{cases}
        [1, 0] & \text{if } u \text{ is the start node} \\
        [0, 1] & \text{if } u \text{ is the goal node}  \\
        [0, 0] & \text{otherwise}
    \end{cases}
    \label{eq:node_features}
\end{equation}

Each edge $e = (u, v)$ is associated with a feature vector 
$\mathbf{f}_e \in R^2$ containing its Euclidean distance $d_e$ and blocking probability $p_e$:

\begin{equation}
    \mathbf{f}_e = [d_e,\ p_e]
    \label{eq:edge_features}
\end{equation}

Since the graph is undirected, each edge is stored once in the forward 
direction. The reverse direction is constructed internally during the forward 
pass to ensure that information propagates symmetrically between both 
endpoints.

\subsubsection{Architecture}

The model consists of three stages: an input encoding stage, a message 
passing stage, and an edge-level decoding stage.

In the \textbf{encoding stage}, node and edge features are independently 
projected into a shared $d$-dimensional latent space through separate linear layers, followed by a ReLU activation.

In the \textbf{message passing stage}, the node embeddings are refined over $L$ \textsc{gat}v2 layers. At each layer, every node aggregates information from its neighbors with attention coefficients conditioned on both the neighboring node embeddings and the connecting edge embeddings. A residual connection and Layer Normalization are applied after each layer to stabilize training across graphs of varying size and topology.

In the \textbf{decoding stage}, the final node embeddings are used to 
produce a value change estimate for each edge. For each edge $(u, v)$, 
an edge representation is constructed by concatenating the embeddings of 
its two endpoints with a symmetric edge embedding, obtained by averaging 
the forward and reverse edge features. This concatenated representation is 
passed through an \textsc{mlp} to produce a scalar output. A Softplus 
activation enforces non-negativity of the predictions, consistent with the 
theoretical property that $v_c^i \geq 0$ under a cost-minimizing value 
function. Certain edges with $p_e \in \{0, 1\}$ are masked to exactly zero, as their blocking state is already known and scouting them yields no 
information gain.

\subsection{Data Generation}

Training data is generated offline through Monte Carlo simulation. For each 
training instance, a random graph is sampled with varying topology, edge 
distances, and blocking probabilities. A single ground robot is assigned a 
start and goal node, and its node features are constructed according to 
Equation~\ref{eq:node_features}. The ground-truth value change 
$v_c^i(b, a_e)$ for each uncertain edge is then estimated by sampling the 
edge's state and computing the resulting change in the robot's optimal path 
cost.

Specifically, for each uncertain edge $e$ with blocking probability $p_e$, 
the ground-truth label is computed as:

\begin{equation}
    v_c^i(b, a_e) = V^i\!\left(b \mid a_e = \text{block}\right) 
                  - V^i\!\left(b \mid a_e = \text{trav}\right)
    \label{eq:gt_label}
\end{equation}

where each value $V^i$ is estimated by averaging the robot's shortest path 
cost over $M$ sampled realizations of the remaining uncertain edges, 
conditioned on the known state of edge $e$. In our experiments, we use 
$M = 1{,}000$ samples per edge, which provides a sufficiently low-variance 
estimate for training purposes.

Since Monte Carlo estimation introduces sampling noise, the raw labels are 
clipped to zero from below before training:

\begin{equation}
    v_c^i \leftarrow \max\!\left(0,\ v_c^i\right)
    \label{eq:clamp}
\end{equation}

This is justified by the theoretical guarantee that $v_c^i \geq 0$ under a 
cost-minimizing value function: knowing the true state of an edge can never 
increase the expected cost of an optimal plan relative to remaining uncertain 
about it. Certain edges with $p_e \in \{0, 1\}$ are assigned a ground-truth 
label of exactly zero, as their states are already known and scouting them 
yields no information.

The dataset is constructed at the single-robot level: each training example corresponds to one robot's start--goal pair on one graph instance. For a scenario with $N$ ground robots, this produces $N$ independent training examples from a single graph, naturally augmenting the dataset by a factor of $N$ without additional simulation cost. The model is trained to minimize a weighted Huber loss over all edges, assigning full weight to uncertain edges and a small regularization weight to certain edges to reinforce the boundary condition at $p_e \in \{0, 1\}$:

\begin{equation}
    \mathcal{L} = \frac{1}{|\mathcal{E}|} \sum_{e \in \mathcal{E}}\, 
                  w_e \cdot L_\delta\!\left(\hat{v}_c^i,\, v_c^i\right)
    \label{eq:loss}
\end{equation}

\begin{equation}
    w_e = 
    \begin{cases}
        1.0 & \text{if } p_e \in (0, 1) \\
        \lambda & \text{otherwise}
    \end{cases}
    \label{eq:weights}
\end{equation}

where $L_\delta$ is the Huber loss with threshold $\delta$, and $\lambda \ll 1$ 
is a small regularization weight applied to certain edges.

\section{Experiments and Results}
\label{sec:Exres}
We evaluated our planning framework, \gls{SAP}, and its variants under travel cost and runtime as performance metrics across three representative environments: \textit{a river-crossing city}, \textit{dense urban town}, and \textit{rural villages}.

\subsection{SAP Variants and Baselines}
We evaluate four planners against the classical \glsentryfull{CTP} as a baseline, which represents the setting without any scouting drone assistance.

\paragraph{\glsentryfull{SAP}} The base variant considers the full drone action set during \gls{MCTS} search. The large branching factor limits tree depth, making it difficult to identify high-quality scouting decisions.

\paragraph{\glsentryfull{SAPD}} This variant prunes the drone action set using the distance-based heuristic (Eq.~\ref{eq:dist_heuristic}), assigning each drone to the \gls{PBP} nearest to the ground team. The reduced branching factor enables deeper \gls{MCTS} search, though the heuristic remains myopic: proximity does not guarantee structural relevance to any ground robot's future plan.

\paragraph{\glsentryfull{SAPIAP}} This variant prunes drone actions by their \gls{IAP} information gain scores (Eq.~\ref{eq:iap}), focusing scouting on edges most likely to alter ground robot behavior. Exact \gls{IAP} evaluation requires Monte Carlo sampling, however, limiting its applicability to real-time settings.

\paragraph{\glsentryfull{SAPL}} This variant replaces exact \gls{IAP} computation with a learned model that predicts information gain directly, preserving decision quality while reducing computational overhead to real-time levels.

\begin{figure}[t]
    \centering
    \setlength{\tabcolsep}{0pt} 
    \begin{tabular}{ccc}
    \includegraphics[width=0.315\linewidth]{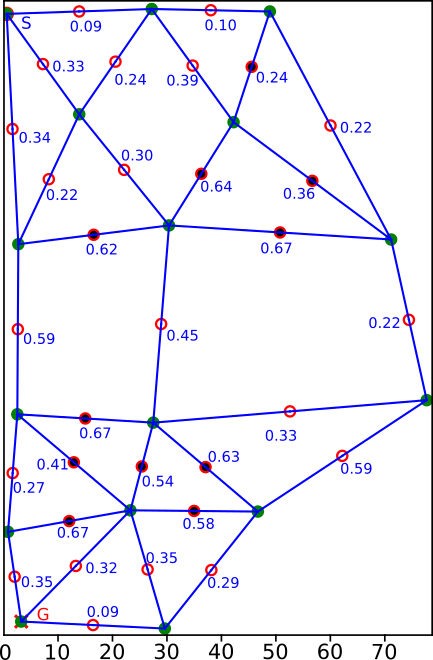} &
    \includegraphics[width=0.288\linewidth]{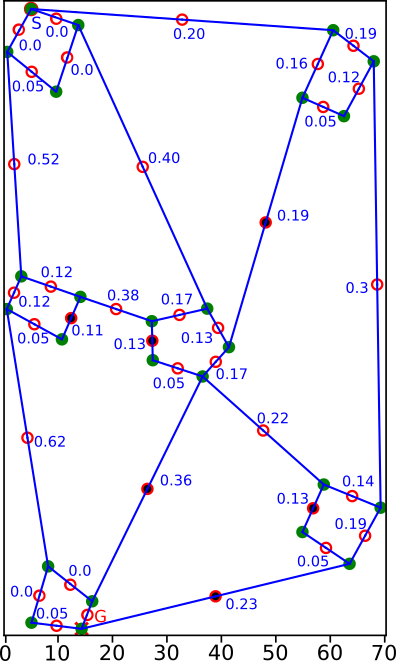} &
    \includegraphics[width=0.401\linewidth]{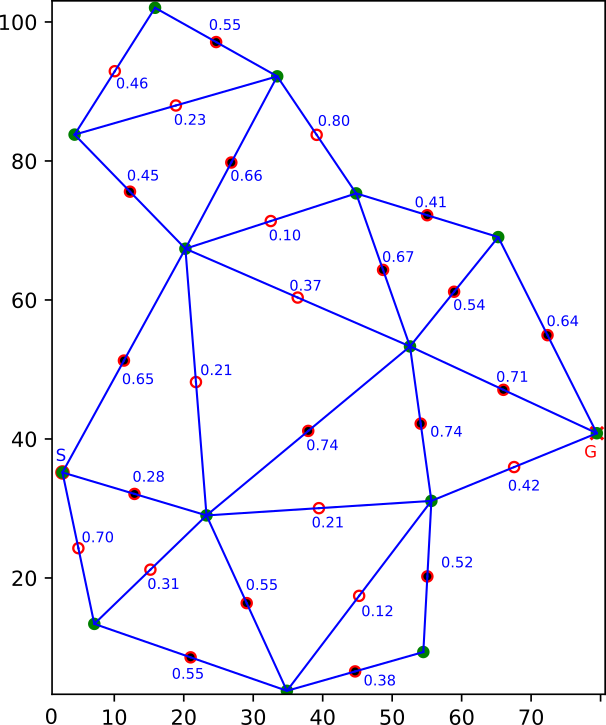} \\
    \footnotesize a) Bridges Graph & \footnotesize b) Islands Graph & \footnotesize c) Random Graph
    \end{tabular}    
    \caption{\textbf{Three graph structures represent three environments}. 
    The \textit{bridges graph} represents \textit{river-crossing cities} with limited bridge connection between two regions.
    The \textit{island graph} captures village-style environments consisting of locally dense street with sparse interconnections between clusters. 
    The \textit{dense-connected graph} models urban areas with high-degree intersections and dense connectivity.
    Each edge has a possible blocking point $\textsc{p}_i$ with a blocking probability $p_i$.}
    \label{fig:graphs}
\end{figure}

\subsection{Environments}

\begin{table*}[tp]
\caption{\textbf{Travel distance and runtime of \textsc{ctp}, \textsc{sap}, \textsc{sap-dap}, \textsc{sap-iap}, and \textsc{sap-liap} across team configurations of up to 3~\glspl{UGV} and 2~\glspl{UAV} on three environments.} All planners use 1,000 \gls{MCTS} rollouts. For \textsc{sap-dap}, \textsc{sap-iap}, and \textsc{sap-liap}, aggressive pruning is applied: each scouting drone is assigned exactly one action per decision step, corresponding to the highest-scoring candidate under the respective pruning strategy. Travel distance and runtime are reported in meters and seconds, respectively. \textsc{ap}-time denotes the time to compute \gls{IAP} action values, while \textit{plan-time} and \textit{No. of steps} denote planning time in each decision step and number of decision steps, respectively. \gls{CTP} serves as the baseline for travel cost comparison.}
\label{tab:exp_table}
\centering
  \footnotesize
  \setlength{\tabcolsep}{0pt} 
  \begin{tabular}{ >{\centering} p{1.3cm}|>{\centering \arraybackslash}p{0.7cm} | >{\centering} p{2.2cm}| >{\centering\arraybackslash}p{1.cm}| >{\centering\arraybackslash}p{1.cm} | >{\centering} p{0.9cm} | >{\centering \arraybackslash} p{2.2cm} | >{\centering\arraybackslash}p{1.cm}| >{\centering\arraybackslash}p{1.cm} | >{\centering \arraybackslash} p{0.9cm} | >{\centering} p{2.2cm}| >{\centering\arraybackslash}p{1.cm}| >{\centering\arraybackslash}p{1.0cm} |>{\centering\arraybackslash} p{0.9cm}}
      \hline
      \multirow{2}{*}{\centering \small Planners} & \multirow{2}{*}{\gls{UGV}}  & \multicolumn{4}{p{5.1cm}|}{\centering \small City with river-crossing \\ (Bridges Graph---Fig.~\ref{fig:graphs}a)} & \multicolumn{4}{p{5.1cm}|}{\centering \small Rural villages \\ (Islands Graph---Fig.~\ref{fig:graphs}b)} & \multicolumn{4}{p{5.1cm}}{\centering \small Dense Urban Town \\(Random Graph---Fig.~\ref{fig:graphs}c)}\\
      \cline{3-14}
      & & {\centering \small Distances} & \small \textsc{ap}-time & \small plan-time & \small No. of steps & {\centering \small Distances} & \small \textsc{ap}-time & \small plan-time & \small No. of steps & {\centering \small Distances} & \small \textsc{ap}-time & \small plan-time & \small No. of steps\\
      \hline
      \multicolumn{14}{p{17.3cm}}{\centering 1 \gls{UGV}} \\
       \hline
       \centering \gls{CTP} & \centering 0 & 384.4 & -- & 2.7 & 15 & 342.3 & -- & 2.6 & 11 & 288.8 & -- & 3.22 & 9\\
       \centering \gls{SAP} & \centering 1 & 364.0 ($\downarrow$ 5.3\%) & -- & 1.6 & 33 & 346.9 ($\downarrow$ -1.3 \%) & -- & 2.2 & 31 & 277.3 ($\downarrow$ 4.0\%) & -- & 2.4 & 27 \\
       \centering \textsc{sap-dap} & \centering 1 & 269.4 ($\downarrow$ 30.0\%) & -- & 2.5 & 31 & 248.6 ($\downarrow$ 27.4\%) & -- & 3.6 & 27 & 212.9 ($\downarrow$ 26.3\%)& -- & 3.7 & 25\\
       \centering \gls{SAPIAP} & \centering 1 & \textbf{239.5} ($\downarrow$ 37.7\%) & 100.0 & 104.0 & 24 & \textbf{214.7} ($\downarrow$ 37.3\%) & 106.4 & 110.3 & 25 & \underline{196.7} ($\downarrow$ 31.9\%) & 130.1 & 133.4 & 17\\
       \centering \textsc{sap-liap} & \centering 1 & \underline{249.2} ($\downarrow$ 35.2\%) & 0.75 & 5.2 & 27 & \underline{223.4}($\downarrow$ 34.7\%) & 0.62 & 5.6 & 29 & \textbf{191.7} ($\downarrow$ 33.6\%) & 0.61 & 4.36 & 22\\
       \hline
       \centering \gls{SAP} & \centering 2 & 268.6 ($\downarrow$ 30.1 \%) & -- & 1.8 & 33 & 241.4 ($\downarrow$ 29.5\%)& -- & 2.4 & 31 & 250.8 ($\downarrow$ 13.1\%)& -- & 2.5 & 28\\
       \centering \textsc{sap-dap} & \centering 2 & 233.1 ($\downarrow$ 39.4\%) & -- & 2.5 & 33 & 224.3 ($\downarrow$ 34.5\%) & -- & 3.2 & 31 & 200.0 ($\downarrow$ 30.7\%)& -- & 3.6 & 28\\
       \centering \gls{SAPIAP} & \centering 2 & \underline{222.8} ($\downarrow$ 42.0\%)& 96.55 & 99.8 & 30 & \textbf{206.3} ($\downarrow$ 39.7\%)& 97.5 & 101.3 & 29 & \underline{196.7} ($\downarrow$ 31.9\%)& 134.5 & 137.5 & 17\\
       \centering \textsc{sap-liap} & \centering 2 & \textbf{221.5} ($\downarrow$ 42.4\%) & 0.53 & 4.1 & 33 & \underline{206.6} ($\downarrow$ 39.6\%) & 0.44 & 5.3 & 31 & \textbf{189.0} ($\downarrow$ 34.6\%) & 0.47 & 4.2 & 28\\
       
       \hline
      \multicolumn{14}{p{17.3cm}}{\centering 2 \glspl{UGV}} \\
       \hline
       \centering \gls{CTP} & \centering 0 & 849.6 & -- & 3.2 & 20 & 784.9 & -- & 3.7 & 19 & 641.1 & -- & 4.6 & 14\\
       \centering \gls{SAP} & \centering 1 & 709.4($\downarrow$ 16.5\%) & -- & 2.3 & 34 & 627.6($\downarrow$ 20.0\%)& -- & 4.1 & 32 & 568.3 ($\downarrow$ 11.4\%)& -- & 3.4 & 28\\
       \centering \textsc{sap-dap} & \centering 1 & 583.8 ($\downarrow$ 31.3\%)& -- & 2.6 & 32 & 452.4 ($\downarrow$ 42.4\%)& -- & 3.1 & 29 & \underline{448.4} ($\downarrow$ 30.1\%)& -- & 3.6 & 26\\
       \centering \gls{SAPIAP} & \centering 1 & \textbf{472.1}($\downarrow$ 44.4\%) & 128.8 & 131.9 & 26 & \textbf{435.1}($\downarrow$ 44.6\%) & 162.0 & 167.5 & 27 &\underline{397.1}($\downarrow$ 38.1\%)& 172.5 & 174.9 & 20\\
       \centering \textsc{sap-liap} & \centering 1 & \underline{479.5} ($\downarrow$ 43.6\%) & 0.49 & 5.1 & 30 & \underline{444.1}($\downarrow$ 43.4\%) & 0.42 & 5.7 & 28 & \textbf{386.1} ($\downarrow$ 39.8\%) & 0.43 & 4.3 & 25 \\
       \hline
       \centering \gls{SAP} & \centering 2 & 541.5($\downarrow$ 36.3\%) & -- & 2.4 & 34 & 494.0($\downarrow$ 37.1\%) & -- & 2.8 & 32 & 514.2 ($\downarrow$ 19.8\%) & -- & 2.6 & 29\\
       \centering \textsc{sap-dap} & \centering 2 & 470.4 ($\downarrow$ 44.6\%)& -- & 2.3 & 34 & \underline{405.3} ($\downarrow$ 48.4\%)& -- & 3.2 & 31 & \underline{383.4} ($\downarrow$ 40.2\%)& -- & 3.4 & 28\\
       \centering \gls{SAPIAP} & \centering 2 & \textbf{438.7} ($\downarrow$ 48.4\%)& 131.6 & 134.7 & 31 & 413.9 ($\downarrow$ 47.3\%)& 168.0 & 173.7 & 31 & 386.4 ($\downarrow$ 39.7\%)& 149.9 & 152.3 & 26 \\
       \centering \textsc{sap-liap} & \centering 2 & \underline{445.2} ($\downarrow$ 47.6\%) & 0.33 & 3.8 & 33 & \textbf{396.8} ($\downarrow$ 49.4\%) & 0.34 & 5.3 & 31 & \textbf{363.2} ($\downarrow$ 43.4\%) & 0.35 & 4.0 & 28 \\
       \hline
      \multicolumn{14}{p{17.3cm}}{\centering 3 \glspl{UGV}} \\
       \hline
       \centering \gls{CTP} & \centering 0 & 1301.5 & -- & 3.3 & 24 & 1190.2 & -- & 3.5 & 23 & 994.8 & -- & 4.4 & 16\\
       \centering \gls{SAP} & \centering 1 & 981.0 ($\downarrow$ 24.6\%)& -- & 2.5 & 35 & 923.2 ($\downarrow$ 22.4\%)& -- & 6.8 & 33 & 810.1 ($\downarrow$ 18.6\%)& -- & 3.8 & 29\\
       \centering \textsc{sap-dap} & \centering 1 & 806.1 ($\downarrow$ 38.1\%)& -- & 2.7 & 33 & 706.0 ($\downarrow$ 40.6\%)& -- & 7.1 & 30 & 667.7 ($\downarrow$ 32.9\%)& -- & 3.7 & 27\\
       \centering \gls{SAPIAP} & \centering 1 & \textbf{719.5} ($\downarrow$ 44.7\%)& 150.5 & 154.1 & 29 &  \textbf{633.3} ($\downarrow$ 46.8\%)& 187.0 & 193.4 & 27 & \textbf{595.6} ($\downarrow$ 40.1\%)& 181.3 & 183.2 & 22\\
       \centering \textsc{sap-liap} & \centering 1 & \underline{729.8} ($\downarrow$ 43.9\%) & 0.34 & 4.8 &32 & \underline{651.6}($\downarrow$ 45.3\%) & 0.29 & 5.5 & 28 & \underline{598.1}($\downarrow$ 39.9\%) & 0.33 & 4.4 & 26 \\
       \hline
       \centering \gls{SAP} & \centering 2 & 786.4 ($\downarrow$ 39.6\%)& -- & 2.4 & 34 & 793.6($\downarrow$ 33.3\%)v& -- & 4.9 & 33 & 762.4 ($\downarrow$ 23.4\%)& -- & 3.2 & 30\\
       \centering \textsc{sap-dap} & \centering 2 & 700.6 ($\downarrow$ 46.2\%)& -- & 2.5 & 34 & 642.1 ($\downarrow$ 46.1\%)& -- & 6.1 & 32 & 575.5 ($\downarrow$ 42.1\%)& -- & 3.4 & 28\\
       \centering \gls{SAPIAP} & \centering 2 & \underline{658.7} ($\downarrow$ 49.4\%)& 158.0 & 161.6 & 33 &  \textbf{614.4} ($\downarrow$ 48.4\%)& 218.0 & 224.9 & 31 & \underline{552.4} ($\downarrow$ 44.5\%)& 156.6 & 159.4 & 23\\
       \centering \textsc{sap-liap} & \centering 2 & \textbf{652.6} ($\downarrow$ 49.9\%) & 0.24 & 3.6 & 34 & \underline{624.2}($\downarrow$ 47.6\%) & 0.29 & 5.9 & 32& \textbf{537.1} ($\downarrow$ 46.0\%) & 0.25 & 3.9 & 29\\
       \hline
  \end{tabular}
\end{table*}

\begin{figure*}[bp]
    \centering
    \includegraphics[width=0.95\linewidth]{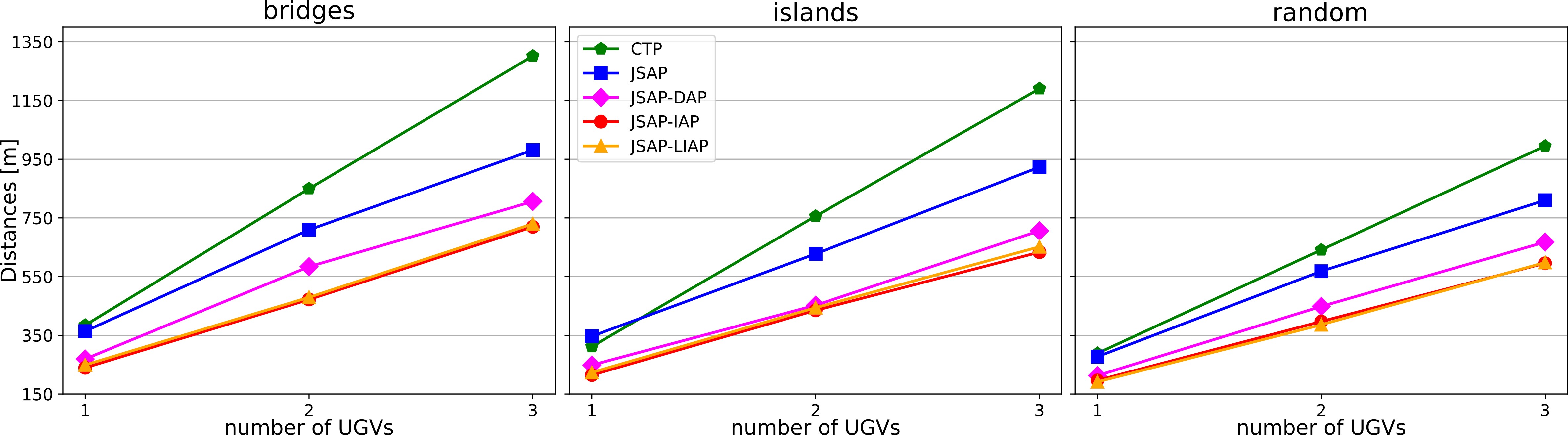}
    \caption{\textbf{Travel Distances of five planners: \textsc{ctp}, \textsc{sap}, \textsc{sap-dap}, \textsc{sap-iap}, and \textsc{sap-liap} with varying number of \glspl{UGV} and 1 \gls{UAV} in three environments.} The performance of \textsc{sap-iap} and \textsc{sap-liap} are comparable in travel cost metric.}
    \label{fig:costs_avg}
\end{figure*}

\paragraph{\bridgesenv} In this environment, there is a river crossing a city, dividing it into two regions (Fig.~\ref{fig:graphs}a). There are several bridges connecting two sides. The environment is modeled as a bridge graph with two group of vertices connecting with each other via some edges.
The starting vertices are on the top of the graph while the goals are on the bottom. The bridge on the shortest path connecting start and goal has the highest blocking probability (in range of ($0.45-0.65$), while the bridge of the medium path has a blocking probability of ($0.35-0.45$), and the bridge on the longest path has blocking probability of ($0.15-0.25$). Other edges have blocking probabilities in range of ($0.1-0.70)$.

\paragraph{\islandenv} This environment consists of several villages that have locally dense streets with sparse long interconnections between them. We model the environment as \textit{islands graph} (Fig.~\ref{fig:graphs}b) in which each island is represented as a group of 4-5 vertices connected by short edges. The blocked probabilities of the local streets are low (in range of ($0.0-0.2$)). The islands are connected by long edges that have high blocked probability of ($0.20-0.65$).

\paragraph{\ranenv} In this environment, the traffic system consists of high-degree intersections and dense connectivity. We model this environment as \textit{random graphs} (Fig.~\ref{fig:graphs}c) with 16 vertices (intersections) that are selected randomly with a minimal distance to a neighbor vertex of 20 m. 
The edges are made by connecting the neighboring vertices using Delaunay triangulation package \cite{2020SciPy-NMeth} with maximal distance of 40 m. We limits the number of edges within (28--30). To model a post-diaster event, we set high blocking probabilities for the edges: 40\% of them having blocking probabilities in range $(0.6-0.8)$, others have blocking values in range of $(0.1-0.6)$. The starting positions are the left vertices on the graph, while the goals are the vertices with the longest distance to the corresponding starting vertices.

\subsection{Results}
We evaluate \gls{SAP}, \gls{SAPD}, \gls{SAPIAP}, and \gls{SAPL} against the \gls{CTP} baseline across three environments with team configurations of up to 3 \glspl{UGV} and 2 \glspl{UAV}, measuring ground robot travel distance and planning runtime. For \gls{SAPD}, \gls{SAPIAP}, and \gls{SAPL}, we apply aggressive pruning in which each scouting drone is assigned exactly one action per decision step---the highest-scoring candidate under the respective pruning strategy. The drone velocity is set to be three times that of the ground robots. 
Results are reported in Table~\ref{tab:exp_table} and Fig.~\ref{fig:costs_avg}.

\begin{figure*}[h]
    \centering
    \setlength{\tabcolsep}{1pt}
    \begin{tabular}{ccc}
        \includegraphics[width=0.326\linewidth]{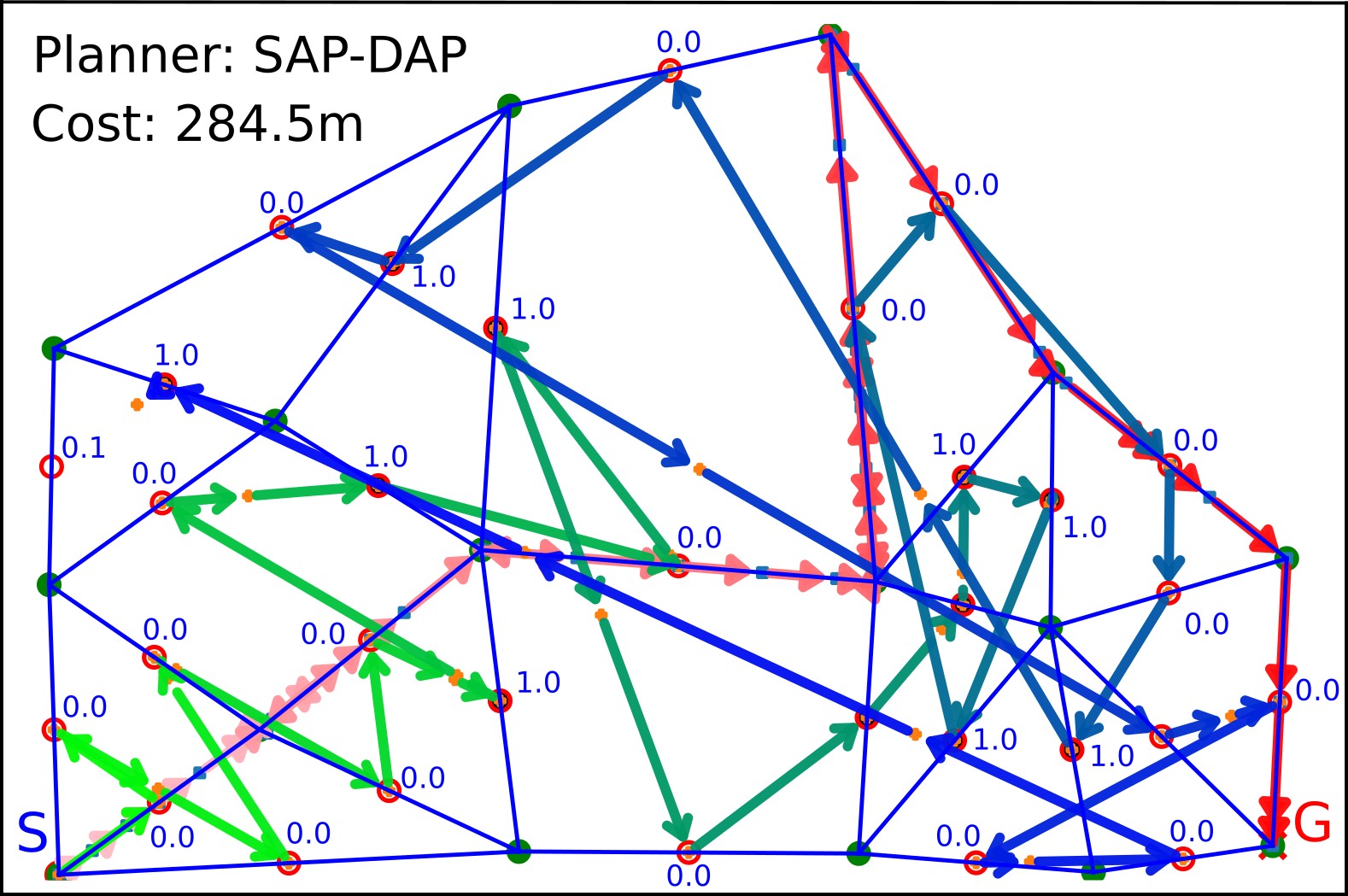} & 
        \includegraphics[width=0.321\linewidth]{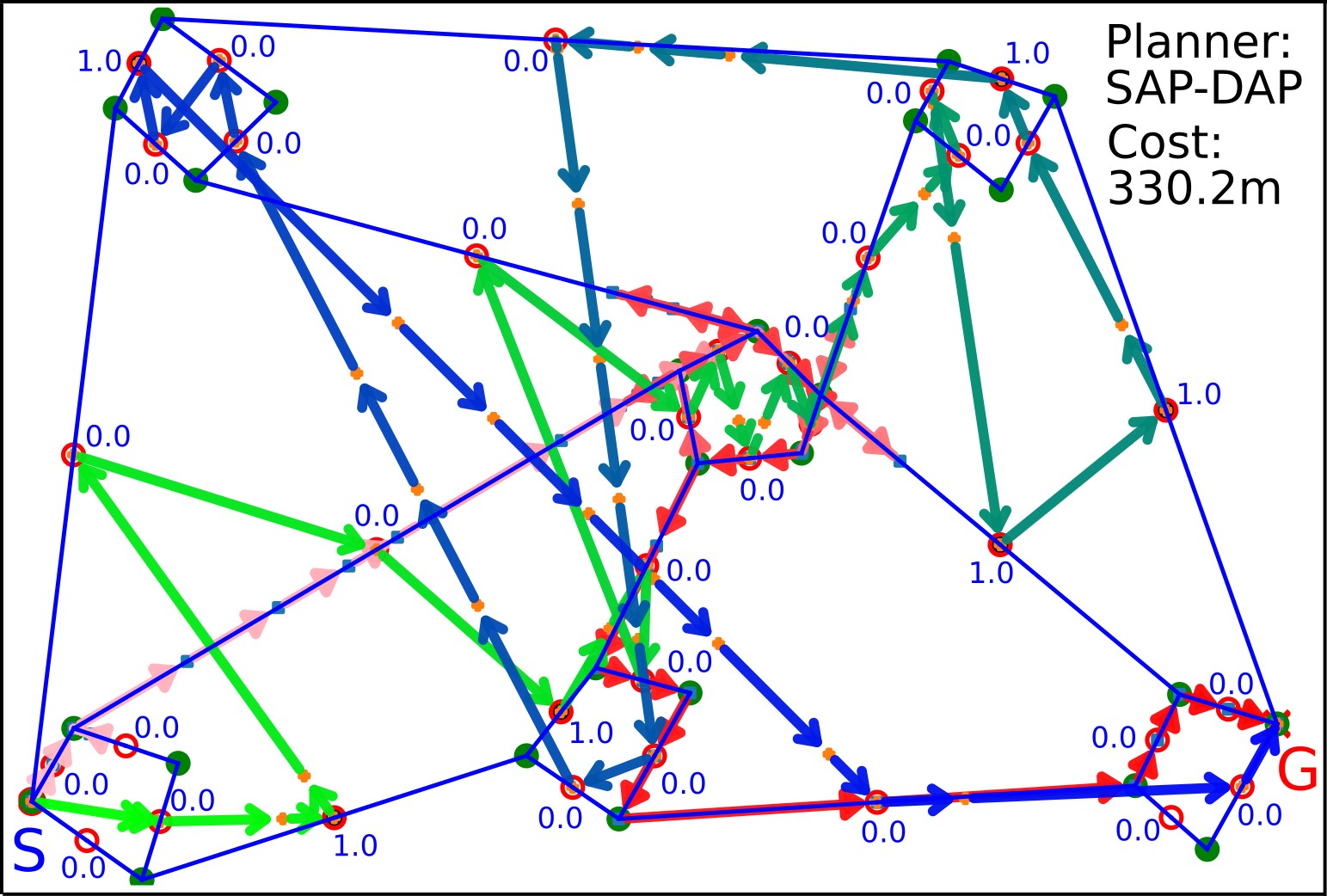} &
        \includegraphics[width=0.345\linewidth]{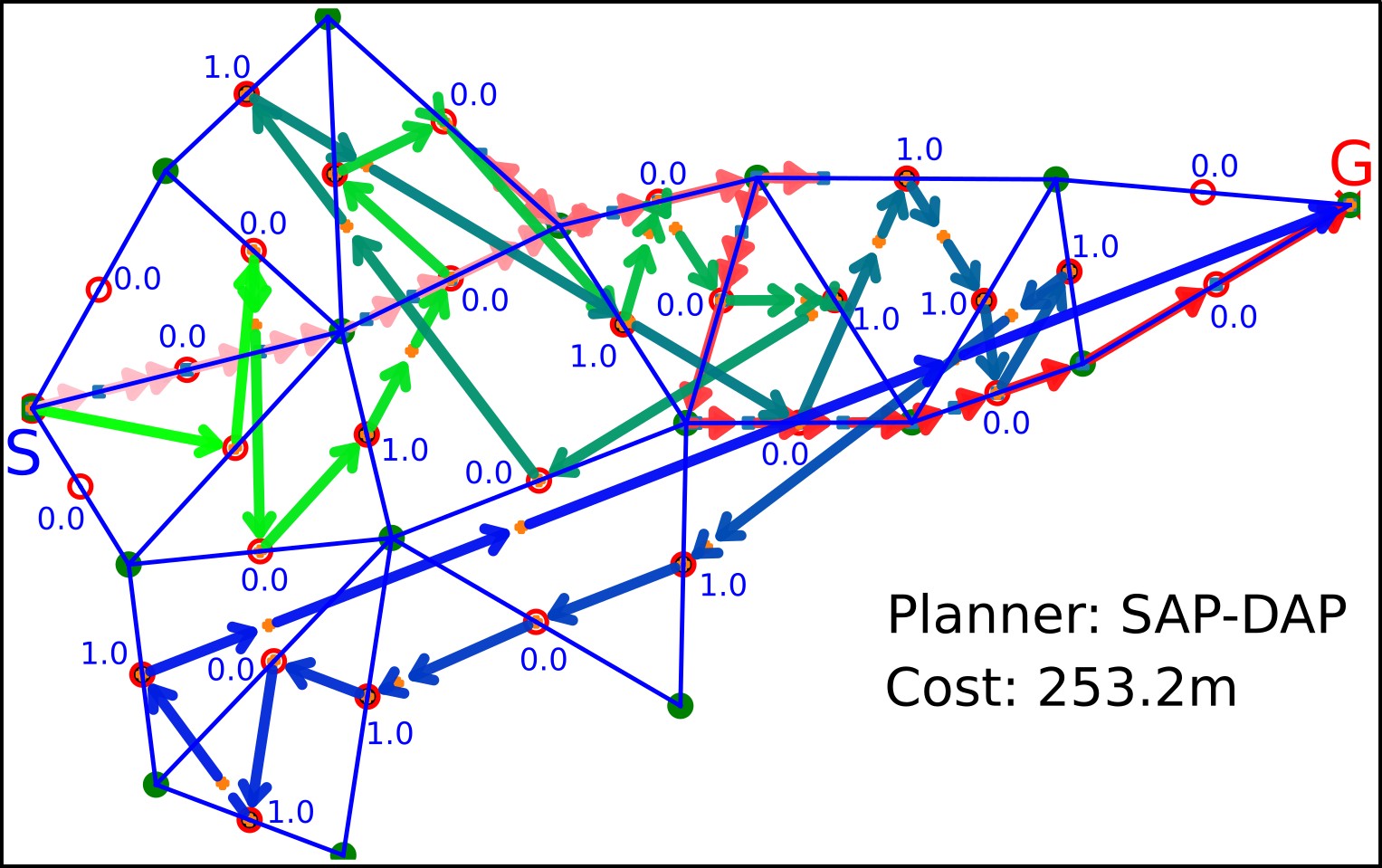}\\
        \includegraphics[width=0.324\linewidth]{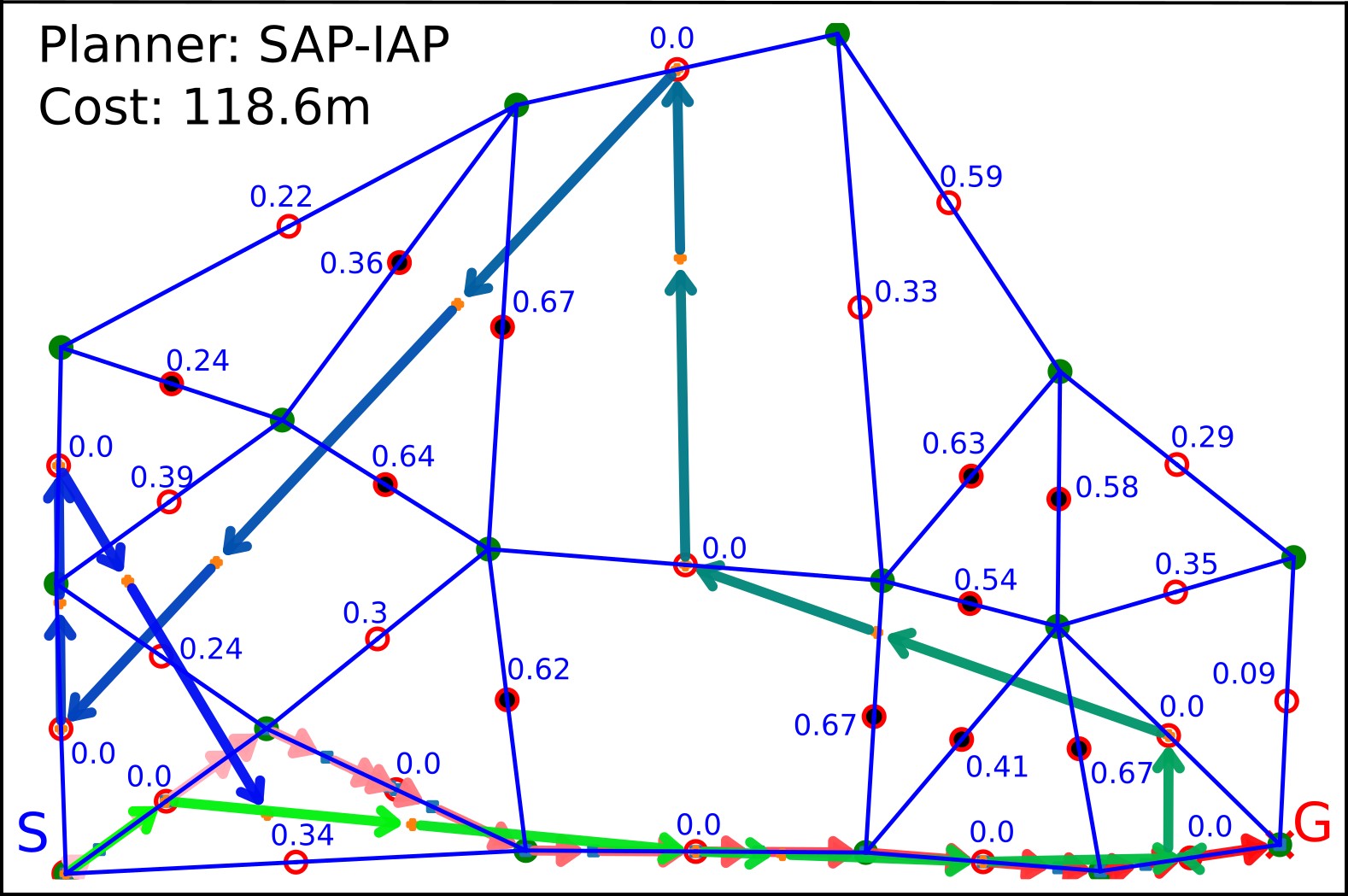} & 
        \includegraphics[width=0.321\linewidth]{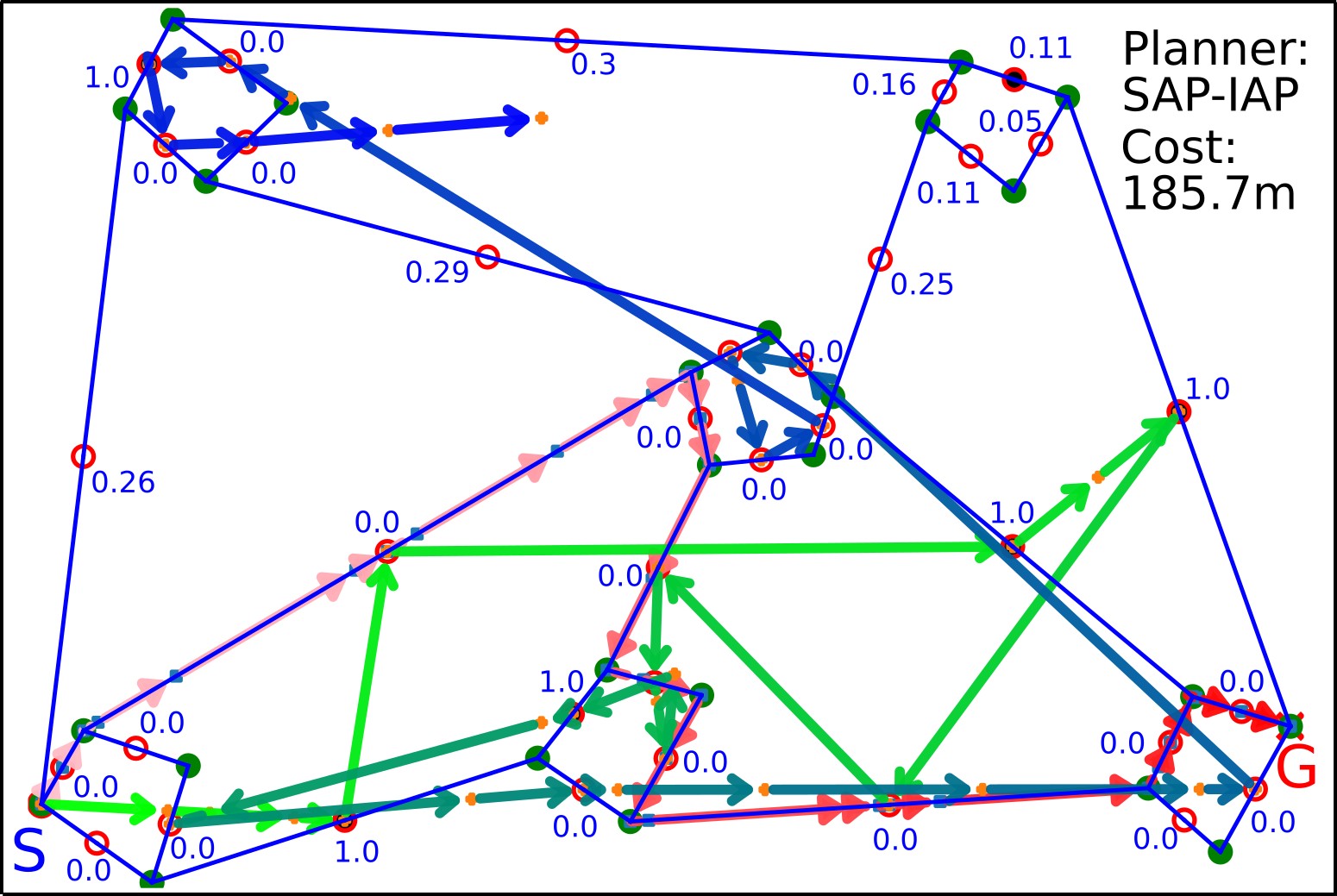} &
        \includegraphics[width=0.345\linewidth]{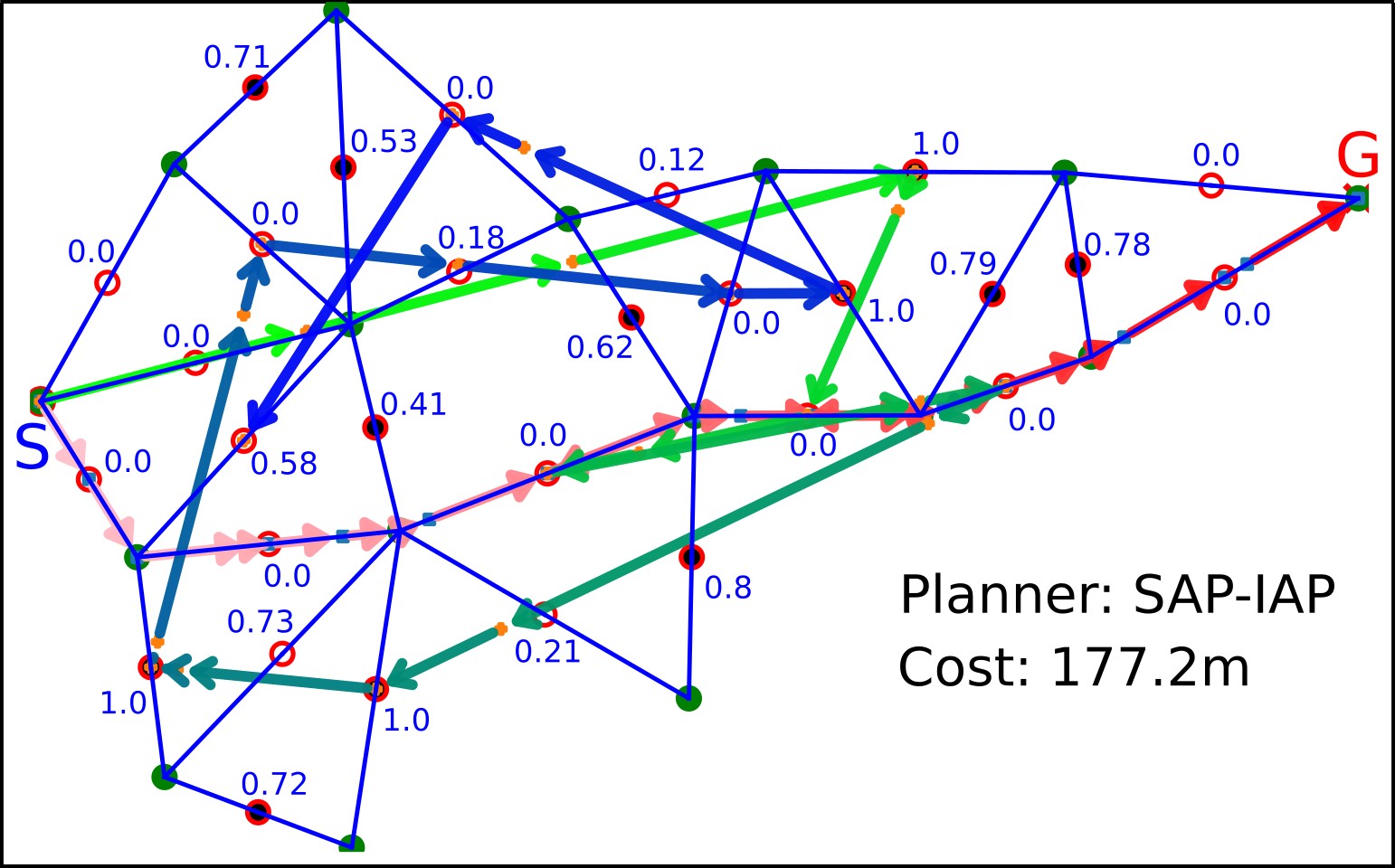}\\
        \includegraphics[width=0.313\linewidth]{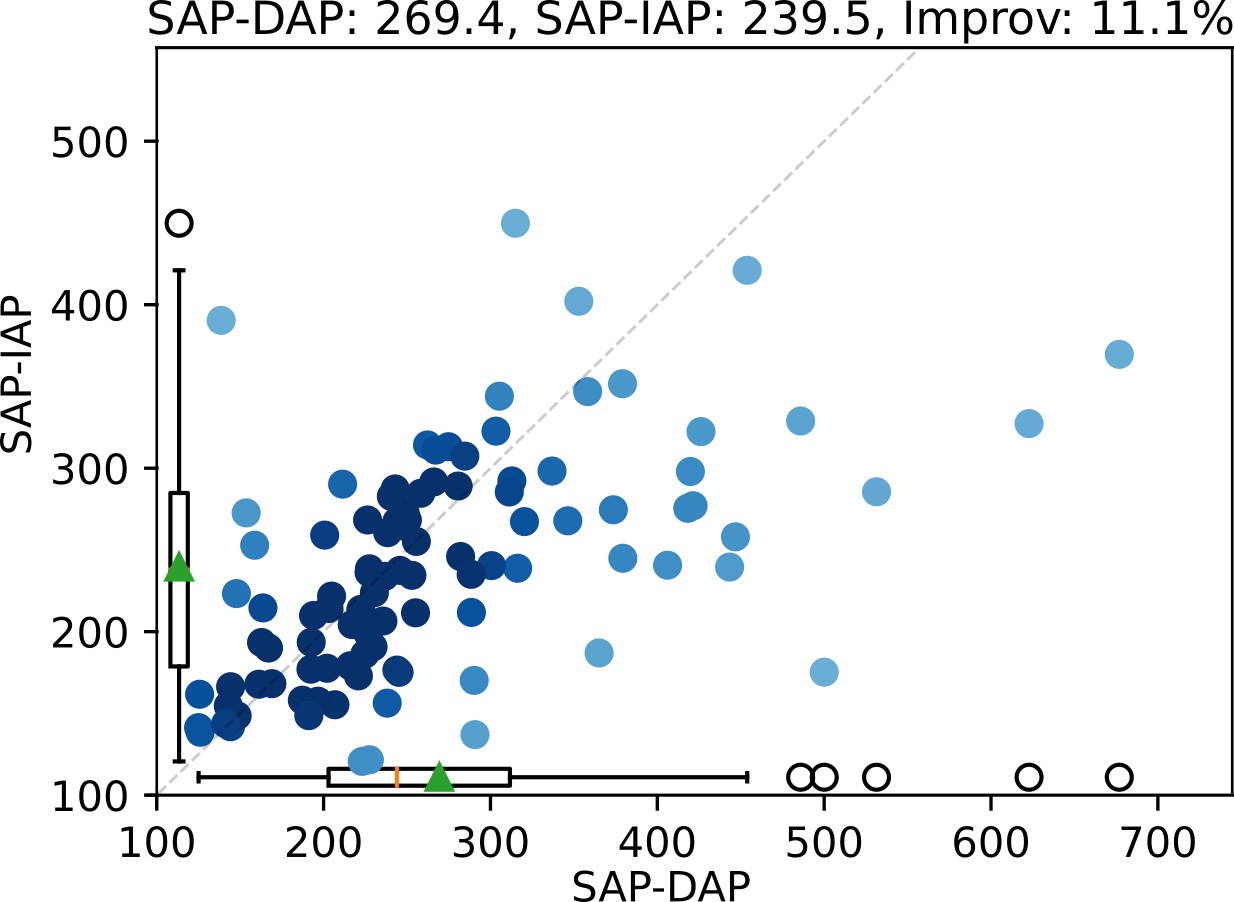} &
        \includegraphics[width=0.3\linewidth]{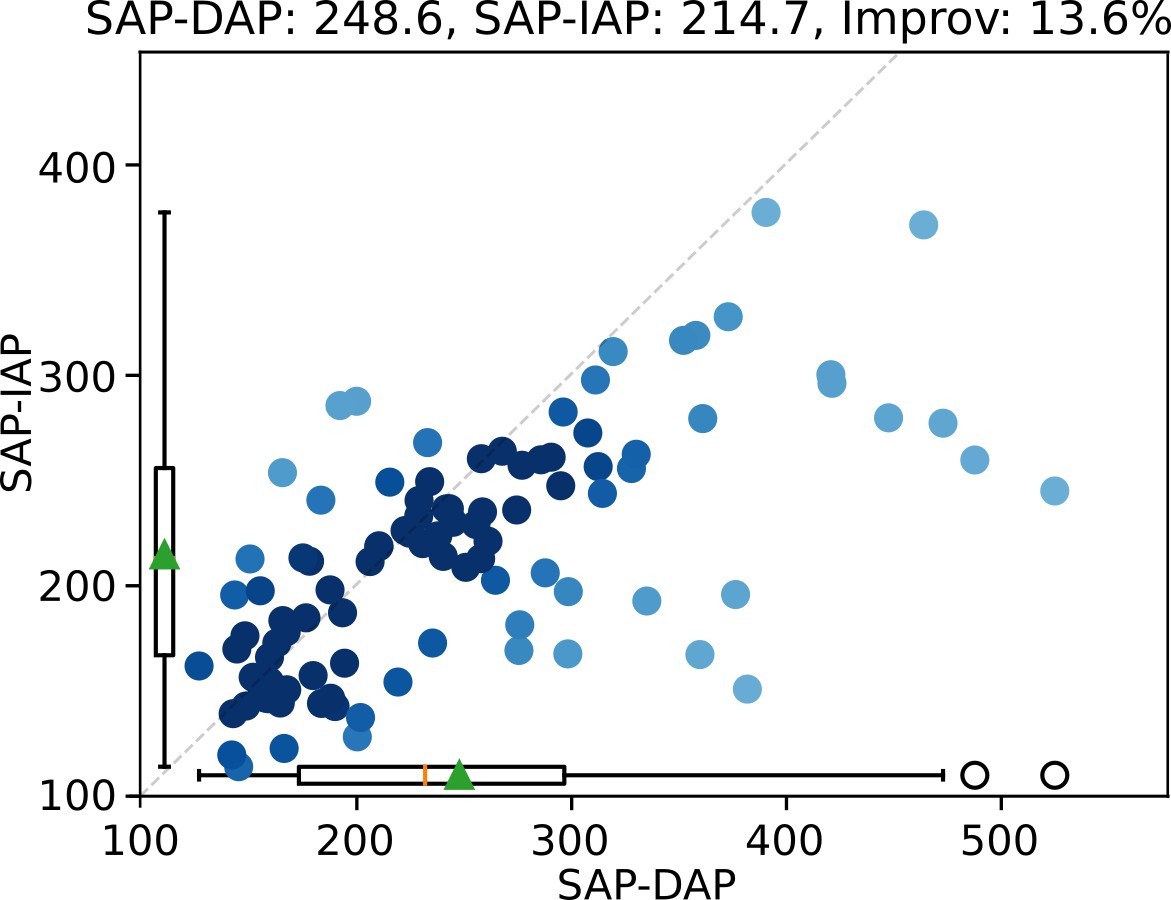} & 
        \includegraphics[width=0.3\linewidth]{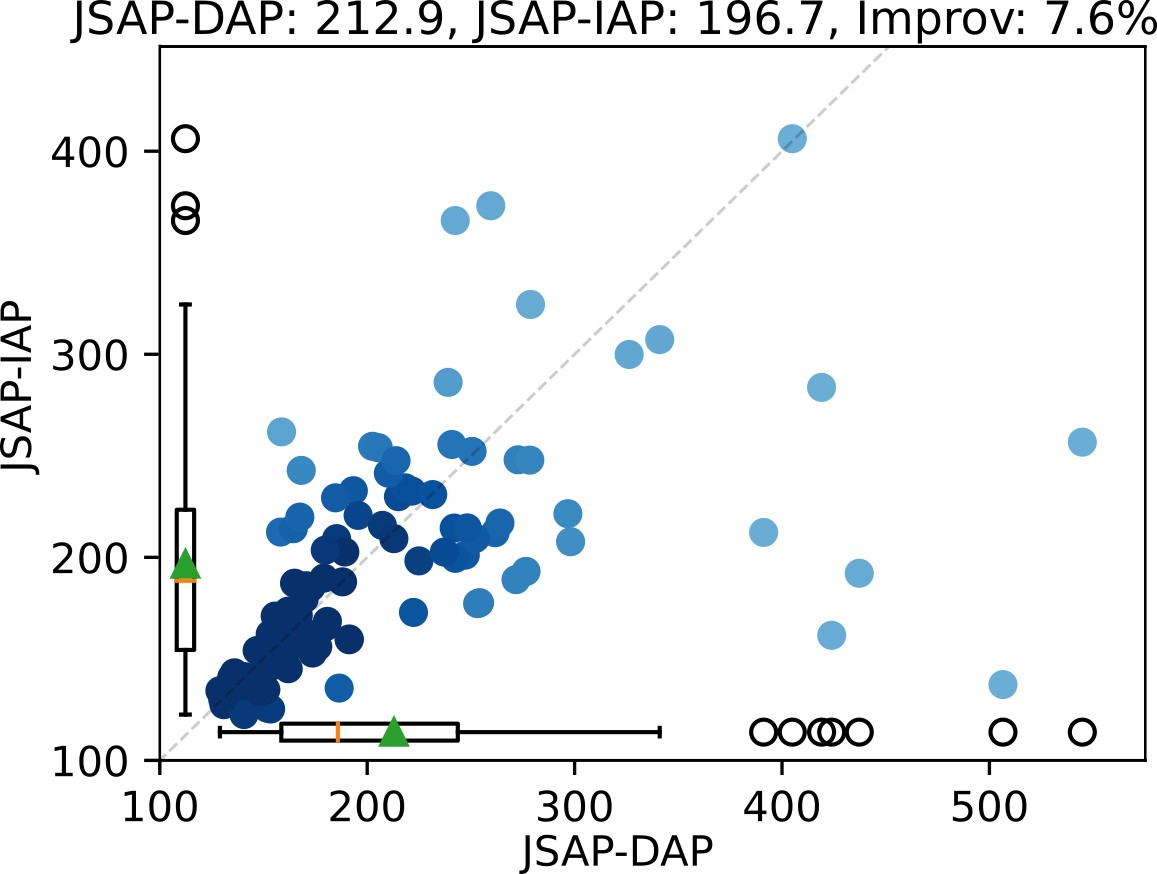}\\
        \footnotesize{a) City with river Crossing(Bridges Graph} &
        \footnotesize{b) Rural Villages (Islands Graph)} &
        \footnotesize{c) Dense-Urban Town (Random Graph)} 
         
    \end{tabular}
    \caption{\textbf{Behavior of \gls{SAPD} and \gls{SAPIAP} in three type of environments}. The purple lines with solid arrow heads represent \gls{UGV}'s paths, while the blue lines with hollow arrow heads demonstrate \gls{UAV}'s paths.}
    \label{fig:behavior_planners}
\end{figure*}

\paragraph{Does scouting guidance reduce ground robot travel cost?}
Scouting drones improve ground robot planning only when their actions are guided by a pruning strategy. Without pruning, \gls{SAP} relies entirely on \gls{MCTS} sampling over the full drone action set, whose exponential branching factor keeps the search tree shallow and scouting decisions uninformed. The result is inconsistent and marginal improvement: a team of 1~\gls{UGV} and 1~\gls{UAV} achieves only 4--5\% travel cost reduction on \bridgesenv{} and \ranenv{} environments, and even slightly underperforms the \gls{CTP} baseline by 1.0\% on \islandenv{} environment.

Guided pruning changes this picture substantially. \gls{SAPD}, which uses the distance-based heuristic to focus scouting on \glspl{PBP} near the ground team, reduces travel cost by 30.0\%, 27.4\%, and 26.3\% on \bridgesenv{}, \islandenv{}, and \ranenv{}, respectively, for a 1~\gls{UGV}--1~\gls{UAV} team. The reduced branching factor enables deeper \gls{MCTS} search, and the locally relevant scouting targets help ground robots avoid committed paths that turn out to be blocked. 
Adding a second drone accelerates environment coverage further: with 2~\glspl{UAV}, all planners achieve at least 13\% travel cost reduction across all environments, as critical edge statuses are resolved earlier in the mission.

\paragraph{Does information gain-based guidance outperform distance-based guidance?}
\gls{SAPIAP} consistently outperforms \gls{SAPD} across all environments and team configurations, confirming that principled information gain scoring identifies more consequential scouting targets than proximity alone. For a 1~\gls{UGV}--1~\gls{UAV} team, \gls{SAPIAP} reduces travel cost by 37.7\%, 37.3\%, and 31.9\% on \bridgesenv{}, \islandenv{}, and \ranenv{}, respectively --- improvements of approximately (8--14) percentage points over \gls{SAPD} in each environment. With 2~\glspl{UAV}, \gls{SAPIAP} achieves 42.0\%, 39.7\%, and 31.9\% reductions on the three environments, maintaining its lead over \gls{SAPD}.

The behavioral difference between the two strategies is illustrated in Fig.~\ref{fig:behavior_planners}. On \bridgesenv{} (Fig.~\ref{fig:behavior_planners}a), the three central bridge edges are structurally critical: their blocking status determines which of the two city regions is accessible, and therefore fundamentally changes the ground robot's route. \gls{IAP} correctly identifies these edges as high-value scouting targets and directs the drone to verify them early, giving the ground robot the information it needs to commit to the shortest viable path. \gls{DAP}, by contrast, guides the drone toward whichever edges are nearest to the \gls{UGV}, regardless of structural relevance. This myopic strategy leaves critical edges unverified until later in the mission, causing the ground robot to commit to a path that may require costly backtracking.

The same pattern appears on \islandenv{} (Fig.~\ref{fig:behavior_planners}b), where \gls{IAP} prioritizes the inter-island connector roads that govern which paths between islands are viable, while \gls{DAP} wastes scouting effort on nearby but structurally unimportant edges. 
On \ranenv{} (Fig.~\ref{fig:behavior_planners}c), the advantage of \gls{IAP} over \gls{DAP} is less pronounced: the denser connectivity means ground robots can reroute through neighboring edges without incurring significant additional travel cost. reducing the penalty for late or misdirected scouting.

\paragraph{Does the GNN preserve solution quality while achieving real-time speed?}
Although \gls{SAPIAP} achieves the best travel cost among all variants in most environments, its exact \gls{IAP} computation requires Monte Carlo sampling at every decision step, resulting in planning times that are too long for real-time deployment (Table~\ref{tab:exp_table}). \gls{SAPL} addresses this by replacing exact sampling with a learned \gls{GNN} model that predicts information gain values directly from the graph structure and belief state. The predicted values are used in place of the sampled ones within the same \gls{IAP} priority formula, leaving the rest of the planning pipeline unchanged.

\gls{SAPL} reduces planning time to within seconds per decision step, making it applicable to real-time settings, while achieving travel costs comparable to \gls{SAPIAP}, even outperforming in \ranenv{} environments. The \gls{SAP} framework also demonstrates scalability to larger ground teams: unlike the approximation-based approach of \citet{stadler2023approximating}, which is restricted to a single ground robot, \gls{SAP} supports up to 3~\glspl{UGV} without modification. Adding more ground robots and scouting drones continues to improve team performance, as broader environment coverage resolves edge uncertainty earlier and enables better coordinated routing across the ground team.

\begin{figure}[b!]
    \centering
    \includegraphics[width=0.9\linewidth]{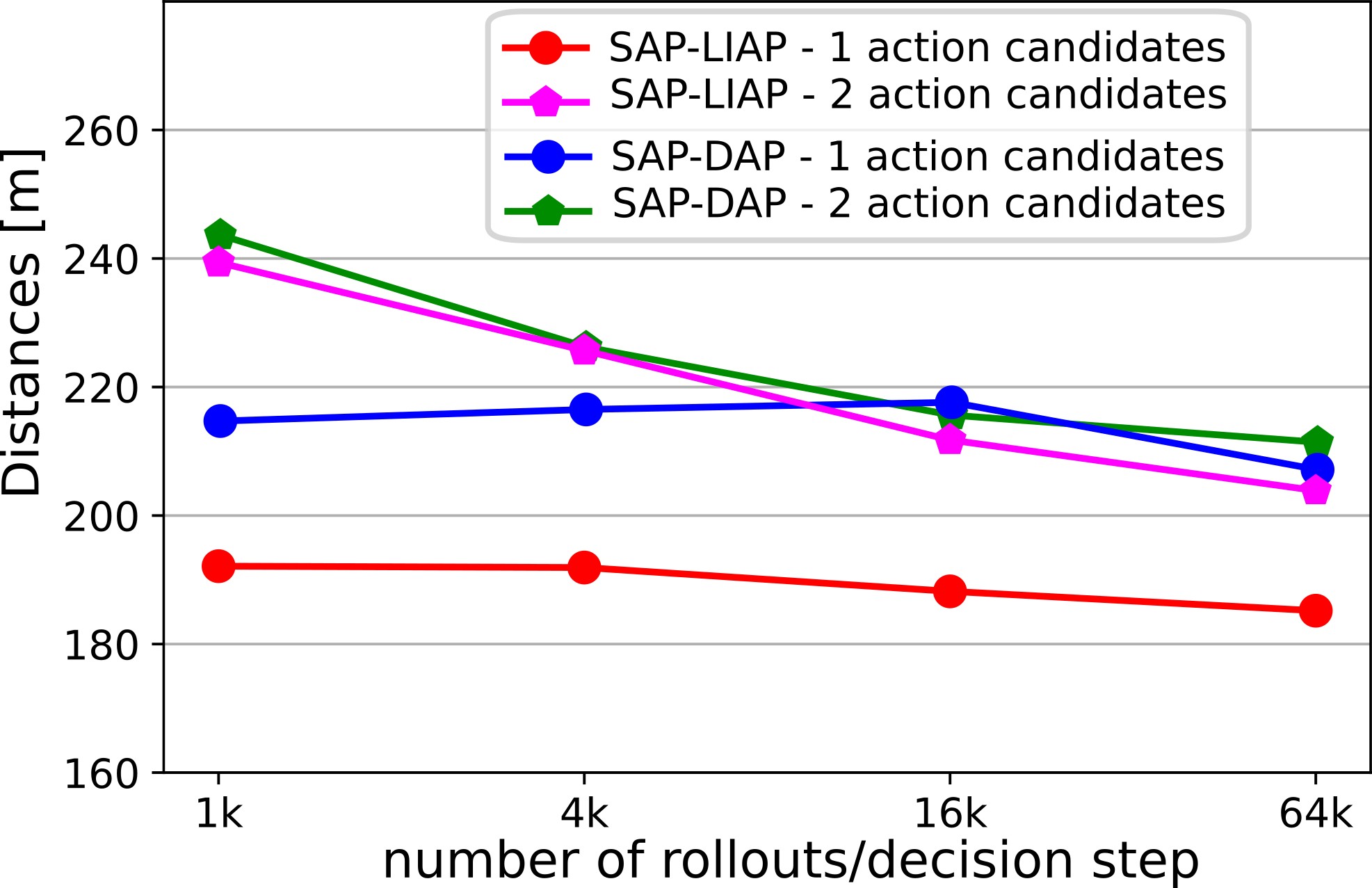}
    \caption{\textbf{Performance of \gls{SAPL} and \gls{SAPD} with one and two action candidates after pruning in \ranenv{} Env.} }
    \label{fig:topK_actions}
\end{figure}

\paragraph{How many drone action candidates should be retained after pruning?}
The \gls{MCTS} search tree grows exponentially with the number of retained drone action candidates, making the choice of pruning aggressiveness critical for both planning quality and tractability. To investigate this, we compare retaining 1 versus 2 action candidates under \gls{SAPL} and \gls{SAPD} while varying the \gls{MCTS} rollout budget from 1000 to 64000, measuring the resulting ground robot travel cost.

As shown in Fig.~\ref{fig:topK_actions}, for the distance-based planner \gls{SAPD}, retaining only a single action candidate at limited rollout budgets (1000 and 4000) improves planner performance. As the rollout budget increases, the performance difference becomes negligible because the planner has sufficient rollouts to effectively sample two action candidates and select a favorable action for execution.

Meanwhile, retaining a single action candidate for the \gls{SAPL} planner consistently outperforms retaining two candidates across all rollout budgets. With 1000 rollouts, the single-candidate setting achieves a travel cost of 192.1m, compared to 239.4m for the two-candidate setting --- a difference of 47.3m. Increasing the rollout budget only marginally reduces this gap: even at 64000 rollouts, the two-candidate setting reaches 203.9m, which remains 18.7m higher than the single-candidate setting at the same budget (185.2m). This persistent performance gap suggests that the additional branching introduced by a second candidate cannot be sufficiently compensated for through increased sampling within a practical rollout budget. Moreover, these results highlight the effectiveness of the \gls{IAP} formulation in identifying actions with the highest information gain. Expanding the action set substantially increases computational effort without yielding corresponding improvements in planner performance.

These results confirm that aggressively pruning to the single top-ranked action is not only computationally advantageous but also produces the best planning performance. Therefore, we adopt this setting across all experiments.


\section{Conclusion} 
\label{sec:Conc}

This paper presents \glsentryfull{SAP}, a heterogeneous planning framework for robot teams operating in partially known environments. \gls{SAP} introduces a unified graph representation and a shared high-level action abstraction that allows both \glspl{UGV} and \glspl{UAV} to operate within a single planning framework, and supports immediate behavior adaptation by ground robots whenever new environmental information becomes available.

The core technical contribution is the \glsentryfull{IAP} formulation, which computes per-edge information gain values to focus scouting drones on actions most likely to alter ground robot behavior. To make this tractable for real-time deployment, we develop a \gls{GNN}-based model that predicts these values directly from graph structure and belief state, reducing planning time from minutes to seconds without sacrificing solution quality. The resulting planner, \gls{SAPIAP}, is scalable to teams of up to 3~\glspl{UGV} and 2~\glspl{UAV}.

Experimentally, our framework consistently outperforms the \gls{CTP} baseline across all environments and team configurations. With a single \gls{UGV} and \gls{UAV}, \gls{SAPIAP} reduces ground robot travel cost by 31.9\%--37.7\% depending on the environment. Travel cost reductions grow further with larger ground teams and additional scouting drones, as broader environment coverage resolves edge uncertainty earlier and enables better coordinated routing. These results confirm that principled, information-gain-guided scouting is both significantly more effective than proximity-based guidance and, with learned value estimation, computationally feasible for real-world deployment.

\bibliographystyle{named}
\bibliography{ijcai26}

\end{document}